%% file: main.tex
\documentclass[10pt,twocolumn,letterpaper]{article}
\usepackage{multirow} 
\usepackage[ruled,vlined]{algorithm2e} 
\usepackage[pagenumbers]{cvpr} 

\input{preamble}
\definecolor{cvprblue}{rgb}{0.21,0.49,0.74}
\usepackage[pagebackref,breaklinks,colorlinks,allcolors=cvprblue]{hyperref}

\def\paperID{4698} 
\def\confName{CVPR}
\def\confYear{2026}

\title{PosA-VLA: Enhancing Action Generation via Pose-Conditioned Anchor Attention}

\author{
Ziwen Li$^{1}$ \and
Xin Wang$^{2}$ \and
Hanlue Zhang$^{1}$ \and
Runnan Chen$^{3}$ \and
Runqi Lin$^{3}$ \and
Xiao He$^{2}$ \and
Han Huang$^{2}$ \and
Yandong Guo$^{2}$ \and
Fakhri Karray$^{1}$ \and
Tongliang Liu$^{1,3}$ \and
Mingming Gong$^{1,4}$ \\
\\
$^{1}$MBZUAI \quad
$^{2}$AI2 Robotics \quad
$^{3}$The University of Sydney \quad
$^{4}$The University of Melbourne
}

\begin{document}
\maketitle
\input{sec/0_abstract}    
\input{sec/1_intro}
\input{sec/2_related}

\input{sec/3_method}

\input{sec/4_experiment}
\input{sec/5_conclusion}


\input{sec/X_suppl}
\clearpage
\newpage
{
    \small
    \bibliographystyle{ieeenat_fullname}
    \bibliography{main}
} 
\end{document}

%% file: sec/0_abstract.tex
\begin{abstract}

The Vision-Language-Action (VLA) models have demonstrated remarkable performance on embodied tasks and shown promising potential for real-world applications. 
However, current VLAs fail to exhibit consistent and precise target-oriented actions, as they often generate redundant motions along trajectories, limiting their applicability in time-sensitive scenarios. 
In this work, we attribute the redundant actions to the spatially uniform perception field of VLAs, which leads them to be distracted by target-irrelevant objects, particularly in complex environments. 
To this end, we propose an efficient \textbf{PosA-VLA} framework, which anchors visual attention via \emph{pose-conditioned supervision}, consistently guiding the model’s perception toward task-relevant regions. 
The pose-conditioned anchor attention mechanism enables the model to better align instruction semantics with actionable visual cues, thereby enhancing action generation precision and efficiency. 
Moreover, our framework is built upon a lightweight architecture and requires no auxiliary perception modules (e.g., segmentation or grounding networks), ensuring efficient inference. 
Extensive experiments verify that our method performs embodied tasks with precise and time-efficient execution across diverse robotic manipulation benchmarks and shows robust generalization in various environments.

\end{abstract}

%% file: sec/1_intro.tex
\section{Introduction}
\label{sec:intro}
\begin{figure}[!t]
    \includegraphics[width=\linewidth]{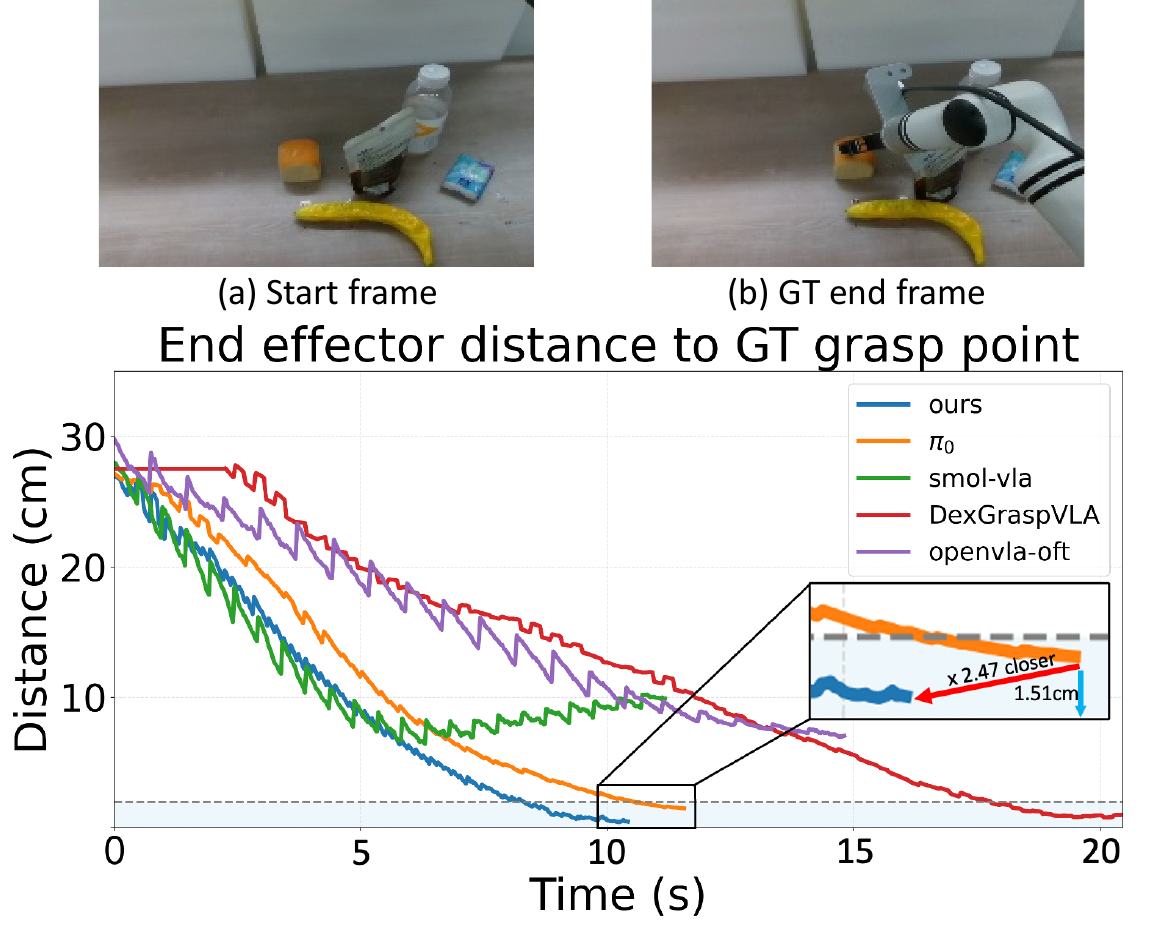}
\vspace{-2em}

    \caption{Quantitative analysis of the grasping task (pick up the bread). 
Top: the initial scene (left) and the ground-truth grasping moment captured from human teleoperation (right). 
Bottom: distance between the robot end-effector and the ground-truth grasp point over time; the light-blue area denotes the successful grasping range. 
Our PosA-VLA reaches the grasping region faster and more accurately, while DexGraspVLA and $\pi_0$ eventually succeed but require longer execution time. 
In contrast, OpenVLA and Smol-VLA fail to reach the successful grasping range.}
    \label{fig:distance}
\vspace{-2 em}
\end{figure}

Despite remarkable progress in artificial intelligence (AI) across various domains, a considerable gap still exists between its computational intelligence and real-world physical interaction. 
Recently, the emergence of vision-language-action (VLA) models~\cite{zhong2025survey,li2024ag2manip,liu2024rdt,yang2023learning,brohan2022rt,zitkovich2023rt,black2410pi0,kim2024openvla,zhong2025dexgraspvla,deng2025graspvla,li2025controlvla,bu2025univla,li2025maniptrans,huang2023embodied,team2024octo} has aimed to bridge this gap by enabling robots to perceive visual scenes, interpret natural language instructions, and execute corresponding manipulation tasks. 
By equipping vision-language understanding models with the ability to act within physical environments, VLAs represent one of the most promising pathways toward achieving embodied AI.

However, these models still struggle to generate consistent and precise actions in real-world applications, which hinders their ability to handle time-sensitive and accuracy-critical tasks. 
In practice, their motion trajectories often exhibit redundant or even unstable behaviors, showing inconsistencies in action generation and insufficient control precision during execution.

As illustrated in Figure~\ref{fig:distance}, current VLA models~\cite{black2410pi0,kim2025fine,shukor2025smolvla,zhong2025dexgraspvla} exhibit considerable fluctuating distance curves between the end-effector and the target point during execution. 
This lack of smoothness primarily arises from frequent action reversals and deviations from the optimal trajectory. 
Such instability often increases the number of execution steps required to complete a single task, thereby reducing overall efficiency. 
More critically, the resulting unstable motion trajectories can lead to more detrimental issues such as unstable contacts, failed grasps, or collisions with nearby objects, as observed in the results of OpenVLA~\cite{kim2025fine} and Smol-VLA~\cite{shukor2025smolvla} in Figure~\ref{fig:distance}.
Together, these issues significantly degrade the models' performance in real-world scenarios, especially when the environment is complex.

We investigate this phenomenon from the perspective of VLA attention and identify that the inconsistency mainly stems from the spatially uniform perception field of existing models. 
We argue that VLAs~\cite{shukor2025smolvla,kim2024openvla,black2410pi0} often rely on a globally \textit{uniform perception field}, lacking explicit mechanisms to attend to both the \textbf{task-relevant regions} and the \textbf{end-effector area}. 
Without such spatial selectivity, the model fails to establish a consistent focus that reflects the true interaction dynamics during manipulation. 
Lacking an internal attention signal conditioned on the robot’s pose, these models tend to \textit{reactively scan the entire scene} rather than proactively concentrating on the most informative areas. 
As a result, they struggle to maintain stable and coherent attention throughout execution, leading to inaccurate or suboptimal actions since their perception is misaligned with regions that truly determine task success. 
This issue becomes particularly evident in cluttered or visually complex environments, where the model’s attention is easily distracted by task-irrelevant objects or background elements, producing unstable motion trajectories and, in severe cases, completely erroneous actions.

To address these challenges, we propose \textbf{PosA-VLA}, a framework that enhances action generation by anchoring visual attention through \emph{pose-conditioned supervision}. 
Specifically, PosA-VLA establishes an explicit link between the robot’s end-effector pose and visual attention, anchoring perception to task-relevant regions throughout the manipulation process. 
It generates two complementary attention anchors: a \textit{task-relevant anchor} obtained specifically at the moments when the end-effector state changes to indicate interaction with regions of interest, and a \textit{end-effector anchor} that tracks the end-effector’s position at each timestep. 
These anchors serve as pose-conditioned spatial priors that transform the continuous 3D interaction space into localized 2D supervision, effectively breaking the model’s spatially uniform perception. 
By dynamically updating the anchors as the robot moves, PosA-VLA maintains a consistent spatial correspondence between the robot’s pose and the visual scene, enabling more stable and precise perception–action coupling. 
This anchoring mechanism allows the model to focus persistently on both the target region and the end-effector’s current position, resulting in smooth, accurate, and efficient manipulation.

Extensive experiments show that PosA-VLA significantly improves task success rates and produces smoother, more stable trajectories while requiring fewer action steps to complete each task compared to baselines. 
Moreover, PosA-VLA is built upon lightweight backbones and operates without auxiliary perception modules (e.g., segmentation or grounding networks), ensuring a compact and efficient architecture. 
Our model also exhibits strong generalization under changes in background, lighting, and distractor configurations, and is capable of handling long-horizon manipulation tasks with consistent performance. 
Notably, PosA-VLA is able to achieve strong performance with only a \textit{small amount of training data}, highlighting its deployment-friendly design and practical applicability.

\noindent\textbf{Our main contributions are summarized as follows:}
\begin{itemize}
    \item \textbf{Empirical analysis of VLA inconsistency.} We identify that the lack of consistent and precise actions in existing Vision-Language-Action (VLA) models originates from their \textit{spatially uniform perception field}, which prevents them from maintaining stable attention on task-relevant regions during manipulation.
    
    \item \textbf{PosA-VLA framework.} We propose PosA-VLA, a VLA architecture that anchors visual attention through pose-conditioned supervision, explicitly linking the robot’s end-effector pose with perception to guide attention toward actionable regions.
    
    \item \textbf{High performance and efficiency.} Extensive experiments on diverse robotic manipulation benchmarks show that PosA-VLA achieves \textit{higher success rates, smoother trajectories, and faster inference} than baseline VLAs, while maintaining robust performance under variations in environment, distractor objects, and lighting conditions.
\end{itemize}

%% file: sec/2_related.tex
\begin{figure*}[t]
    \centering
    \includegraphics[width=0.9\linewidth]{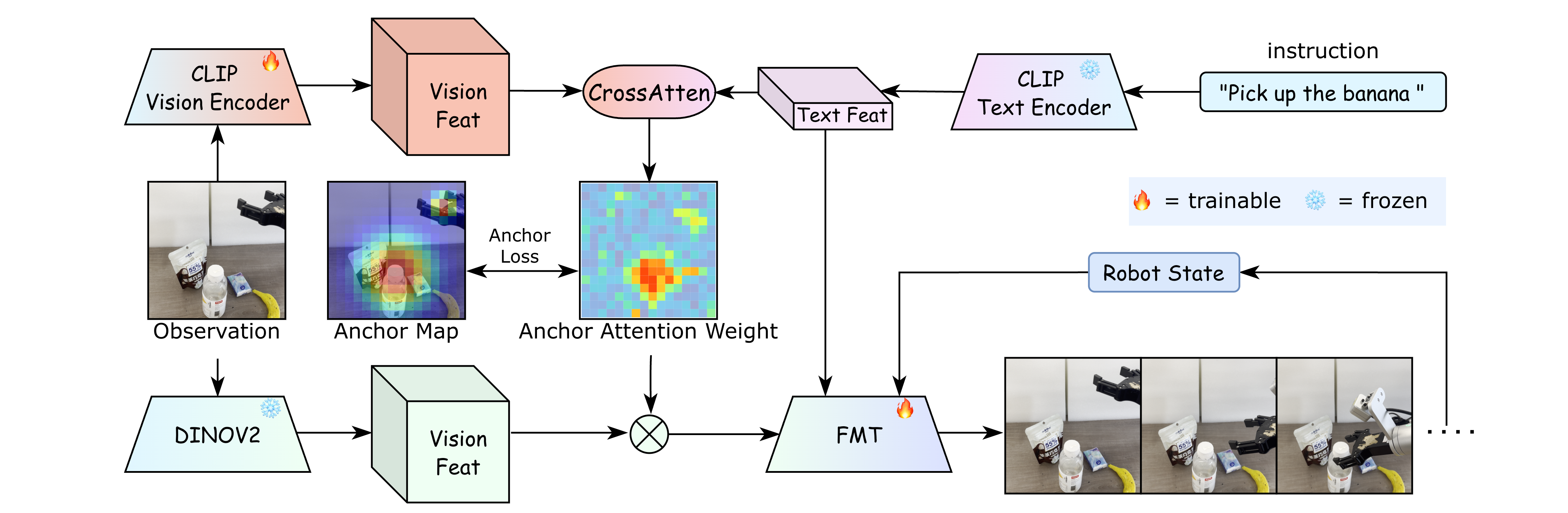}
    \caption{Overview of the proposed \textbf{PosA-VLA} framework. 
A CLIP text encoder extracts the textual feature, while a CLIP image encoder produces patch-wise visual features from head and wrist cameras. 
These features are fused through a cross-attention module to generate anchor attention weights, which are supervised by the proposed anchor loss using the ground-truth pose-conditioned anchor maps.
The anchor attention weights are then applied to DINOv2 image features via element-wise multiplication to obtain refined visual representations. 
Finally, the refined visual features, together with the text feature and the robot state feature, are fed into the Flow Matching Transformer (FMT) to predict the continuous action sequence.
}
\label{fig:framework}
\vspace{-10 pt}
\end{figure*}
\section{Related Work}

\subsection{Vision-Language-Action Models}
Recent advances in vision-language-action (VLA) learning have enabled robots to interpret natural language instructions and execute grounded control behaviors. 
Early frameworks explored direct end-to-end training that maps raw visual observations and language instructions to motor commands through behavior cloning. Representative works~\cite{bharadhwaj2024roboagent,zawalski2024robotic} demonstrated the feasibility of grounding natural language into low-level manipulation by jointly encoding multimodal inputs within a unified policy network. 
Recent works such as RT-1~\cite{brohan2022rt} and RT-2~\cite{zitkovich2023rt} scale robot control via large multimodal transformers pretrained on web-scale image-text-action data~\cite{chen2023pali,driess2023palm}. 
This progress has also been driven by the emergence of large-scale robot demonstration datasets and benchmarks~\cite{o2024open,dasari2019robonet,walke2023bridgedata,ramos2021rlds,khazatsky2024droid,liu2023libero}, 
which provide diverse real-world manipulation trajectories and language-conditioned control data for model pretraining and evaluation. 
More recent architectures, including $\pi_0$~\cite{black2410pi0},$\pi_{0.5}$~\cite{intelligence2025pi_}, OpenVLA~\cite{kim2024openvla,kim2025fine} and others~\cite{li2023vision,bu2025univla,ye2024latent,stone2023open} employ large-scale multimodal~\cite{zhai2023sigmoid,team2024gemma,touvron2023llama} that achieve strong generalization across diverse embodied tasks but come with significant computational overhead and slow inference. 
SmolVLA~\cite{shukor2025smolvla} presents a lightweight alternative~\cite{marafioti2025smolvlm} that improves efficiency while maintaining reasonable performance. 
However, all these models share a common limitation: they still require extensive finetuning on downstream manipulation datasets to achieve task-specific proficiency. 
Moreover, their perception remains largely \textit{spatially uniform}, lacking explicit mechanisms to maintain task-relevant attention, which often leads to redundant or imprecise trajectories in cluttered or dynamic environments.

\subsection{Spatial Grounding in Embodied AI}
Spatial grounding is essential for connecting abstract linguistic references with concrete 3D environments. 
Existing methods~\cite{wang2019deep,devin2018deep} commonly address this via object segmentation or grounding networks such as Grounding DINO~\cite{liu2024grounding,ren2024grounding} and SAM~\cite{kirillov2023segment}, or by leveraging large pretrained models~\cite{bai2023qwen,wang2024qwen2,cai2024internlm2} as in ~\cite{zhong2025dexgraspvla,li2025controlvla,deng2025graspvla}. 
Although effective, these approaches rely heavily on segmentation accuracy and thus on external models, which limit their robustness in dynamic or unseen scenes. 
In contrast, our method introduces a pose anchor that directly projects the end-effector position into the image space, achieving fully self-contained visual attention guidance.

\subsection{Affordance Inference and Manipulation Priors}
A closely related line of work leverages models to infer \textit{task affordances}~\cite{chu2019learning}—i.e., where and how to interact with the environment—to improve manipulation performance in novel settings. 
Recent approaches~\cite{locatello2020object,li2024manipllm,huang2024rekep,huang2024copa,huang2023voxposer,su2025resem3d,ahn2022can,zhu2023learning,tang2023graspgpt} employ large vision-language or segmentation models to detect affordance regions by correlating textual queries with visual cues. 
These methods effectively use pretrained models to localize actionable areas, guiding either downstream motion planning or a learned policy head. 
While powerful, such affordance inference frameworks rely heavily on the zero-shot generalization of large foundation models and often require costly visual-grounding pipelines or post-hoc region proposals. 
Moreover, their affordance maps are computed independently of the robot’s current pose, leading to coarse and sometimes inconsistent spatial grounding during closed-loop control. 
In contrast, our pose-conditioned attention anchor directly encodes geometric relationships between the end-effector and target object, providing a continuous, task-aware attention signal that evolves with the robot’s motion and requires no external foundation models or segmentation annotations.

%% file: sec/3_method.tex
\label{sec:intro}

\section{Methods}
In this section, we present the proposed \textbf{PosA-VLA} framework, which integrates pose-conditioned anchor attention with efficient action generation. 
As illustrated in Figure~\ref{fig:framework}, our method first formulates the language-conditioned manipulation problem (Section~\ref{subsec: problem}), 
then constructs pose-conditioned attention anchors to provide explicit spatial supervision (Section~\ref{subsec:focus region}), 
introduces a pose-conditioned anchor loss to enhance attention consistency and multimodal alignment (Section~\ref{subsec:focus loss}), 
and finally employs a flow matching transformer to generate precise and temporally coherent action sequences (Section~\ref{subsec:action generation}).
\subsection{Problem Formulation}
\label{subsec: problem}
We focus on language-conditioned robotic manipulation, where the model learns to generate precise and consistent control actions given multimodal sensory inputs. 
At each timestep $t$, the robot observes the environment through a head camera $\mathbf{I}^h_t$, a wrist camera $\mathbf{I}^w_t$, its proprioceptive state $\mathbf{s}_t$, and receives a textual instruction $\mathbf{x}$. 
The objective is to predict the corresponding low-level control command $\hat{\mathbf{a}}_t$ that drives the robot to execute the task specified by $\mathbf{x}$.

Formally, the Vision-Language-Action (VLA) model learns a policy 
\begin{equation}
\pi_\theta: (\mathbf{I}^h_t, \mathbf{I}^w_t, \mathbf{s}_t, \mathbf{x}) \rightarrow \hat{\mathbf{a}}_t,
\label{eq:policy}
\end{equation}
where $\theta$ denotes the learnable parameters of the network.

\begin{figure}[!t]
    \centering
    \includegraphics[width=0.9\linewidth]{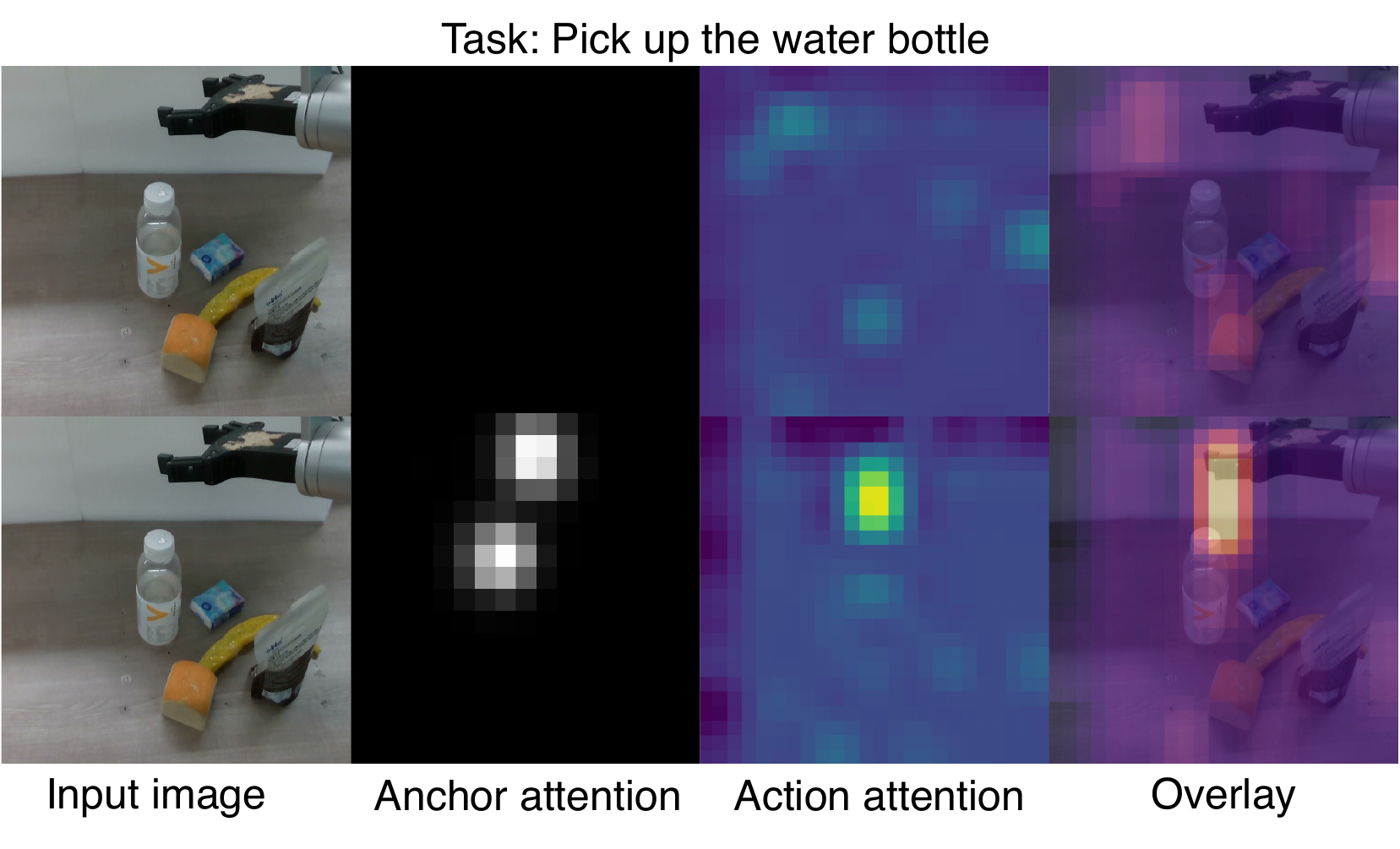}
    \vspace{-0.6 em}
    \caption{Visualization of attention behaviors with and without our anchor supervision. 
Columns (left to right): input image, pose-anchored attention weight ($\mathbf{M}_t$), average cross-attention of the last-layer heads in the action transformer, and overlay of the action attention on the original image. 
Top: baseline without anchor loss; bottom: our PosA-VLA with anchor loss, which produces sharper, more localized, and task-centered attention.}
\vspace{-1.5 em}
    \label{fig:attention}
\end{figure}

\subsection{Pose-Conditioned Anchor Generation}
\label{subsec:focus region}
During preliminary experiments, we observed that the last-layer cross-attention (averaged across all heads) of the action transformer exhibits a spatially uniform activation when trained \emph{without} our anchor loss~\ref{subsec:focus loss}. 
As shown in Figure~\ref{fig:attention}, this baseline spreads attention almost evenly across the scene rather than concentrating on the manipulated object, degrading the accuracy of the generated actions, see Section~\ref{subsec:abl module}.
To provide explicit spatial grounding, we construct \emph{pose-conditioned attention anchors} from the robot’s end-effector trajectories toward target region recorded in the collected demonstrations.

Specifically, we first employ CLIP~\cite{radford2021learning} text and image encoders to obtain language features $\mathbf{f}_x$ and visual features $\mathbf{F}_I$ from both cameras. 
Here, $\mathbf{f}_x \in \mathbb{R}^{d}$ denotes the global text embedding extracted from the CLIP text encoder, 
while $\mathbf{F}_I \in \mathbb{R}^{H\times W\times d}$ corresponds to the patch-wise visual embeddings from the \textit{last hidden state} of the CLIP image encoder.
These representations are fused through a cross-attention module to produce an task-relevant attention weight $\mathbf{M}_t^{\text{task}}\!\in\![0,1]^{H\times W}$. We then use an end-effector query text embedding $\mathbf{f}_e$ (e.g., ``gripper'') to produce an end-effector attention weight $\mathbf{M}_t^{\text{end}}\!\in\![0,1]^{H\times W}$. 
Two maps are stacked along the channel dimension to obtain the pose-anchored attention weight
\begin{equation}
\mathbf{M}_t = \big[\,\mathbf{M}_t^{\text{task}},\, \mathbf{M}_t^{\text{end}}\,\big] \in [0,1]^{H \times W \times 2},
\label{eq:anchor_weight}
\end{equation}
which jointly reflects where the instruction points to and where the end-effector should be attended.

We then identify all timesteps where the end-effector state changes—i.e., when the gripper closes or opens, which indicate that the end-effector is interacting with a task-relevant region. 
For each timestep, we extract the end-effector’s 3D position from its pose $\mathbf{p}_t = [x_t, y_t, z_t, R_t]$ and project it onto both camera views using the camera–end-effector transformation matrix to obtain 2D coordinates $(u^h_t, v^h_t)$ and $(u^w_t, v^w_t)$, where $(u^h_t, v^h_t)$ and $(u^w_t, v^w_t)$ denote the pixel coordinates of the end-effector projection in the head and wrist camera frames, respectively.
A Gaussian map centered at these coordinates defines the task-relevant anchor map $\mathbf{F}_{f}^{\text{task}}$:
\begin{equation}
\mathbf{F}_{f}^{\text{task}}(i,j) = 
\exp\!\left(-\frac{(i-u_t)^2 + (j-v_t)^2}{2\sigma_{\text{task}}^2}\right),
\label{eq:task_anchor}
\end{equation}
where $\sigma_{\text{task}}$ controls the spatial spread of the region of interest.

In addition, we argue that the model should also explicitly attend to the end-effector’s position at every timestep to maintain spatial awareness throughout the manipulation. 
Therefore, we project the end-effector’s 3D position at each frame onto the 2D image plane and generate a smaller Gaussian map $\mathbf{F}_{f}^{\text{end}}$ with a narrower variance $\sigma_{\text{end}}{<}\sigma_{\text{task}}$:
\begin{equation}
\mathbf{F}_{f}^{\text{end}}(i,j) = 
\exp\!\left(-\frac{(i-u_t)^2 + (j-v_t)^2}{2\sigma_{\text{end}}^2}\right).
\label{eq:task_anchor}
\end{equation}
$\mathbf{F}_{f}^{\text{task}}$ and $\mathbf{F}_{f}^{\text{end}}$ serve as spatial supervision signals for $\mathbf{M}_t$ separately, encouraging the model to focus on the end-effector’s current location as well as the task-relevant interaction regions. 
This dual supervision enables the learned attention to capture not only where to act but also how the end-effector moves during manipulation, resulting in more grounded and stable visual focus.

\subsection{Pose-Conditioned Anchor Loss}
\label{subsec:focus loss}
We jointly supervise spatial attention prediction and language–vision alignment via a combination of a spatial classification loss and a batch-wise contrastive loss (pseudocode provided in the Appendix).
\paragraph{Spatial Attention Loss}
The main supervision is imposed on the predicted pose-anchored attention weight $\mathbf{M}_t$, which is trained against the combined supervision $\mathbf{F}_f=[\mathbf{F}_f^{\text{task}},\mathbf{F}_f^{\text{end}}]$.
Following our implementation, we employ a Focal Loss~\cite{lin2017focal} formulation to handle the imbalance between foreground and background embeddings:
\begin{equation}
\mathcal{L}_{f} = \text{FocalLoss}(\mathbf{M}_t, \mathbf{F}_f).
\label{eq:focal_loss}
\end{equation}

\paragraph{Batch-wise Contrastive Loss.}
To further enhance consistency and discriminability across samples, we introduce a batch-wise contrastive loss operating on paired text–image embeddings.
At each training step, the model receives textual embeddings $\mathbf{f}_x$ and visual embeddings $\mathbf{F}_I$, together with the task and end-effector level anchor maps $\mathbf{F}_f^{\text{task}}$ and $\mathbf{F}_f^{\text{end}}$ serving as spatial supervision signals.
Positive indices are selected when the activation exceeds a threshold $\tau_{\text{pos}}{=}0.7$:
\begin{equation}
\Omega^{+} = \{(i,j) \mid \mathbf{F}_f(i,j) > \tau_{\text{pos}}\},
\label{eq:positive_indices}
\end{equation}
where $\mathbf{F}_f$ can be either $\mathbf{F}_f^{\text{task}}$ or $\mathbf{F}_f^{\text{end}}$ depending on the supervision type. 
The image embeddings corresponding to these indices in $\mathbf{F}_I$ are denoted as $\mathbf{v}_{ij}$, which represent the visual features of the selected positive patches. 
Accordingly, the end-effector embedding $\mathbf{v}_{\text{end}}$ and the object embedding $\mathbf{v}_{\text{obj}}$ are both drawn from the subset of $\mathbf{F}_I$ indexed by $\Omega^{+}$.
Depending on whether both end-effector and task-relevant signals are available, we form fused embeddings by concatenating their corresponding visual and textual features. 
A signal is considered available only if the length of $\Omega^{+}$ is greater than zero; otherwise, the corresponding embedding is replaced with zero-padding to maintain a consistent feature dimension.
For instance, when both cues are valid, we concatenate the language embedding $\mathbf{f}_x$ and $\mathbf{f}_e$ with the object visual and end-effector embeddings:
\begin{equation}
\mathbf{z} = [\,\mathbf{f};\, \mathbf{v}_{\text{end}};\, \mathbf{v}_{\text{obj}}\,],
\label{eq:z_concat}
\end{equation}
whereas for task-only supervision, we concatenate $\mathbf{f}_x$ and $\mathbf{v}_{\text{obj}}$ only.  
All fused embeddings in the batch are collected into a set $\mathcal{Z}=\{\mathbf{z}_n\}_{n=1}^{N}$, each associated with a caption identifier $g_n$ based on the textual instruction.

We compute pairwise similarities with a learnable temperature $\tau$:
\begin{equation}
s_{mn} = \frac{\mathbf{z}_m^\top \mathbf{z}_n}{\tau}, \qquad 
\tau = \exp(\theta_{\tau}) \text{ is learned.}
\label{eq:similarity}
\end{equation}

Binary targets are defined by caption group consistency,
\begin{equation}
y_{mn} = \mathbf{1}[g_m = g_n], \quad m \neq n,
\label{eq:binary_target}
\end{equation}
and the loss is implemented as a binary cross-entropy over all off-diagonal pairs:
\begin{equation}
\begin{aligned}
\mathcal{L}_{\text{c}}
=\frac{1}{|\mathcal{P}|}\!\!\sum_{(m,n)\in\mathcal{P}}
\Big(\!-\;y_{mn}\log\sigma(s_{mn}) \\
-\left(1-y_{mn}\right)\log\!\big(1-\sigma(s_{mn})\big)\Big).
\end{aligned}
\end{equation}
where $\mathcal{P}=\{(m,n)\mid m\neq n\}$ and $\sigma(\cdot)$ denotes the sigmoid.  
This batch-wise formulation aligns embeddings from the same instruction while separating those from different tasks, encouraging the network to attend consistently to both $\mathbf{F}_f^{\text{end}}$ and $\mathbf{F}_f^{\text{task}}$ regions.

The total loss is
\begin{equation}
\mathcal{L}_{\text{anchor}} = \alpha\,\mathcal{L}_{\text{f}} + (1-\alpha)\,\mathcal{L}_{\text{c}},
\label{eq:anchor_loss}
\end{equation}
where $\alpha$ balances spatial supervision and batch-wise contrast (we set $\alpha{=}0.5$ by default).

\subsection{Action Generation}
\label{subsec:action generation}
The attention-refined visual features are obtained by element-wise multiplication between the anchor weight and DINOv2~\cite{oquab2023dinov2} image features:
\begin{equation}
\mathbf{F}_{v}^{\text{ref}} = \mathbf{M}_t \odot \mathbf{F}_{\text{DINO}},
\label{eq:refined_features}
\end{equation}
which effectively highlights the most relevant regions for manipulation. 
The refined features, together with the textual embedding $\mathbf{f}_x$ and robot state $\mathbf{s}_t$, are fused into a multimodal observation representation $\tilde{\mathbf{z}}_{t}^{\text{obs}}$ used for conditional action generation.

\begin{figure*}[!t]
    \includegraphics[width=\linewidth]{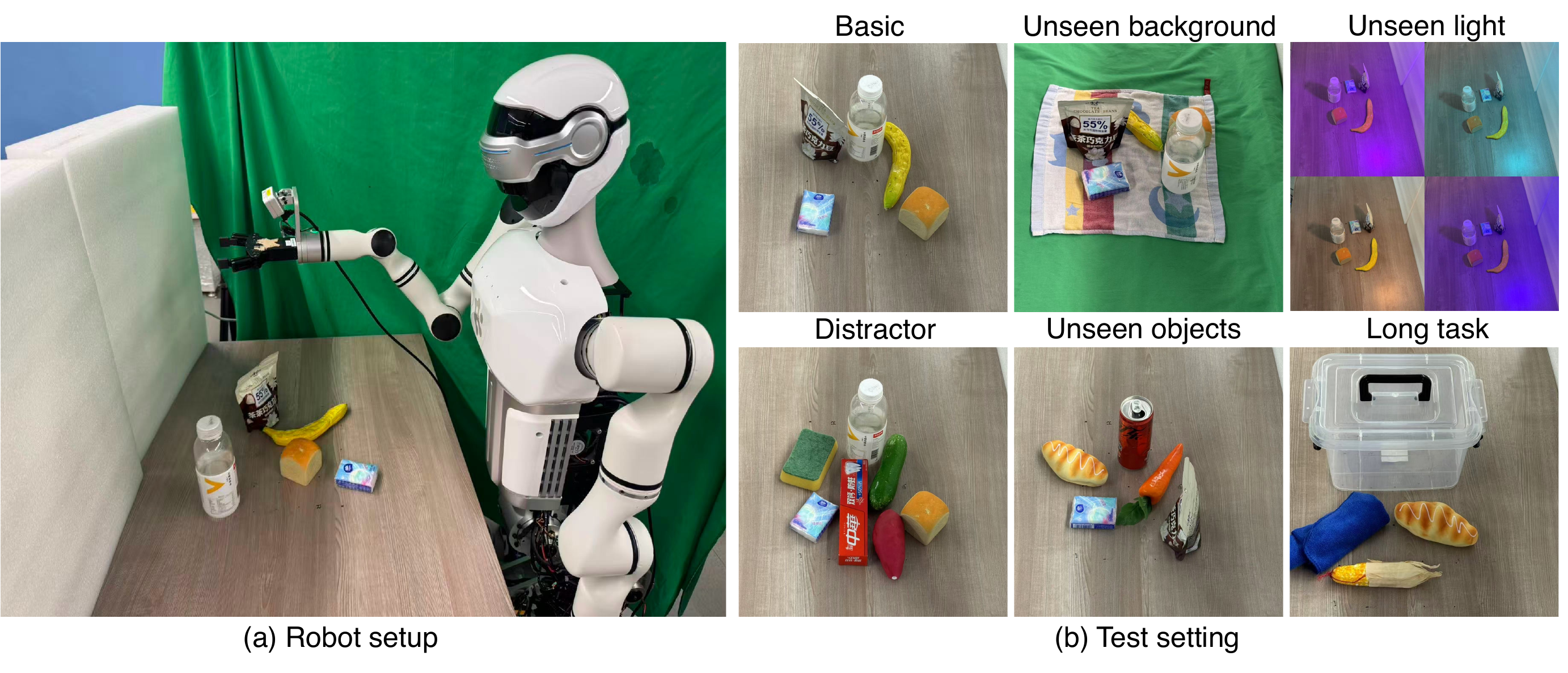}
    \caption{Experimental setup and evaluation environments. 
Left: the robotic platform. 
Right: representative testing environments used in our experiments, including 
\textit{Basic}, \textit{Unseen Background}, \textit{Unseen Lighting}, \textit{Distractor Objects}, \textit{Unseen Objects}, and \textit{Long-horizon Task}.}
    \label{fig:robot}
    \vspace{-1.em}
\end{figure*}

For action prediction, we adopt a \textbf{Flow Matching Transformer (FMT)} that learns a continuous transport field between a simple base distribution and the target action distribution, following recent advances in flow matching for policy learning~\cite{lipman2022flow,black2410pi0,shukor2025smolvla}. 
At each timestep $t$, the next $H$ low-level actions are grouped into a short-horizon \textbf{action chunk} $\mathbf{A}_t = [\mathbf{a}_t, \mathbf{a}_{t+1}, \ldots, \mathbf{a}_{t+H-1}]$. 
The model samples an intermediate time $\tau \in [0,1]$ and constructs an interpolated point along the trajectory between the Gaussian prior $\mathbf{z}_0$ and the target $\mathbf{A}_t$:
\begin{equation}
\mathbf{x}_\tau = (1-\tau)\mathbf{z}_0 + \tau \mathbf{A}_t,
\label{eq:interpolation}
\end{equation}
The FMT then predicts the instantaneous velocity (flow) that transports $\mathbf{x}_\tau$ toward the target distribution:
\begin{equation}
\hat{\mathbf{v}}_\theta = f_\theta(\mathbf{x}_\tau, \tau, \tilde{\mathbf{z}}_{t}^{\text{obs}}),
\label{eq:velocity}
\end{equation}
and is optimized via a velocity-matching objective:
\begin{equation}
\mathcal{L}_{\text{action}} = \mathbb{E}_{\tau}\!\left[\|\hat{\mathbf{v}}_\theta - (\mathbf{A}_t - \mathbf{z}_0)\|_2^2\right],
\label{eq:action_loss}
\end{equation}
This training process directly learns a smooth, deterministic mapping that transforms noise samples into task-conditioned action sequences, avoiding iterative denoising and improving computational efficiency.

During inference, actions are generated by integrating the learned velocity field over time using an ODE solver:

\begin{equation}
\frac{d\mathbf{x}_\tau}{d\tau} = f_\theta(\mathbf{x}_\tau, \tau, \tilde{\mathbf{z}}_{t}^{\text{obs}}), 
\qquad \mathbf{x}_0 \sim \mathcal{N}(0, I),
\label{eq:ode}
\end{equation}
which yields the final clean action chunk $\hat{\mathbf{A}}_t = \mathbf{x}_1$. 
The total training objective combines the flow matching action loss with the anchor supervision term:
\begin{equation}
\mathcal{L}_{\text{total}} = \mathcal{L}_{\text{action}} + \lambda \mathcal{L}_{\text{anchor}},
\label{eq:total_loss}
\end{equation}
where $\lambda$ balances attention alignment and control accuracy.

%% file: sec/4_experiment.tex
\section{Experiments}
\label{sec:exp}

In this section, we evaluate the effectiveness of our method, including experimental setups (Section \ref{sec: setup}), our implementation details (Section \ref{sec: implementation}, performance evaluations (Section \ref{sec: result}), and ablation studies (Section \ref{sec: ablation}).

\subsection{Experimental setups}
\label{sec: setup}
Our experiments are conducted on an \textbf{AlphaBot~1s} (Figure~\ref{fig:robot}(a)) robotic platform equipped with a 7-DoF manipulator, a head-mounted camera, and a wrist-mounted camera. 
We manually collected a dataset comprising multiple grasping tasks involving distinct objects, with 200 demonstrations per object to train our model. 
We compared \textbf{PosA-VLA} against several representative Vision-Language-Action (VLA) baselines, including \textbf{$\pi_0$}~\cite{black2410pi0}, \textbf{$\pi_{0.5}$}~\cite{intelligence2025pi_}, OpenVLA-OFT~\cite{kim2025fine}, Smol-VLA~\cite{shukor2025smolvla}, and DexGraspVLA~\cite{zhong2025dexgraspvla}. 

During testing, object placements were randomly generated to ensure unseen configurations while maintaining visibility and avoiding occlusions (details provided in the Appendix). 
We additionally evaluated model generalization under variations in background, lighting, distractor objects, and unseen target objects.
For the standard grasping tasks, the evaluation metric is defined as a successful grasp followed by lifting the object to a stable position in midair without slippage or collision. 

To further examine robustness and efficiency, we conducted experiments under limited training data and designed a long-horizon task where the robot opens a box lid and places a designated object inside. 
For this long-horizon task, we report both the step-wise success rates of each sub-action (i.e., open and put down the lid, pick up the object, place the object into the box) and the overall task success rate, which is counted only when all stages are successfully completed. 
Finally, we compared the action efficiency and inference speed across all methods.
Simulation experiments and qualitative visualizations are provided in the Appendix.

\begin{table*}[t]
\centering
\caption{Comparison of grasp success rates (\%) under different testing conditions. 
PosA-VLA achieves the highest average success rate.}
\setlength{\tabcolsep}{4.5pt}
\label{tab:grasp_success}
\begin{tabular}{lcccccc}
\toprule
\textbf{Method} & \textbf{Basic} & \textbf{Unseen Background} & \textbf{Unseen Lights} & \textbf{Distractor Objects} & \textbf{Unseen Objects} & \textbf{Average} \\ 
\midrule
$\pi_0$~\cite{black2410pi0}             & 43.1 & 30.2 & 23.6 & 27.9 & 33.4 & 31.6 \\ 
OpenVLA-OFT~\cite{kim2025fine}          & 21.6 & 17.2 & \,\,6.1 & 18.4 & 10.3 & 14.7 \\
Smol-VLA~\cite{shukor2025smolvla}             & 20.3 & 15.8 & \,\,7.9 & 16.5 & \,\,8.7 & 13.8 \\
DexGraspVLA~\cite{zhong2025dexgraspvla}          & 57.2 & 51.2 & 45.0 & 51.1 & \textbf{48.1} & 50.5 \\
$\pi_{0.5}$(LoRA)~\cite{intelligence2025pi_}          & 38.5 & 31.2 & 22.4 & 25.3 & 30.8 & 29.6 \\
$\pi_{0.5}$ (Full)~\cite{intelligence2025pi_}          & 60.8 & 52.5 & 41.3 & 48.6 & 38.9 & 48.4 \\
\midrule
\textbf{PosA-VLA} & \textbf{74.9} & \textbf{57.2} & \textbf{47.8} & \textbf{56.1} & 40.7 & \textbf{55.3} \\
\bottomrule
\end{tabular}
\vspace{-1em}
\end{table*}

\begin{table}[t]
\centering
\caption{Success rates (\%) on the long-horizon box task. 
The task includes three stages: open and put down the lid, pick up the object, and put the object in.}
\label{tab:long_task}
\resizebox{\columnwidth}{!}{
\begin{tabular}{lccc}
\toprule
\textbf{Method} & \textbf{Open\&Put down} & \textbf{Pick up} & \textbf{Overall}  \\
\midrule
$\pi_0$             & 97.2 & 50.0 & 42.6  \\
OpenVLA-OFT          & 73.2 & 15.8 & 13.0  \\
Smol-VLA             & 88.5 & 26.4 &  19.1  \\
\midrule
\textbf{PosA-VLA} & \textbf{97.2} & \textbf{63.0} & \textbf{61.1} \\
\bottomrule
\end{tabular}}
\vspace{-10 pt}
\end{table}

\begin{table}[t]
\centering
\caption{
Comparison of training and inference efficiency. 
We report the total training time (GPU hours) and three inference metrics: 
average time per action (ms), total action steps per task, and total execution time including all system overheads (s).
}
\setlength{\tabcolsep}{1.5pt}
\label{tab:efficiency}
\begin{tabular}{l|c|ccc}
\toprule
\multirow{2}{*}{\textbf{Method}} & \multicolumn{1}{c|}{\textbf{Training}} & \multicolumn{3}{c}{\textbf{Inference}} \\
\cmidrule(lr){2-2} \cmidrule(lr){3-5}
 & \textbf{Time (h)} & \textbf{Avg. Time (ms)} & \textbf{Steps} & \textbf{Total (s)} \\
\midrule
$\pi_0$ (JAX)        & 25 & 27.0  & 563  & 15.2  \\
OpenVLA-OFT          & 104  & 38.9  & 558  & 21.7 \\
Smol-VLA             &  \textbf{19} & \textbf{23.1}  & 624  & 14.4  \\
DexGraspVLA          & 90 & 49.6  & 577  & 28.6 \\
\midrule
\textbf{PosA-VLA} & 20 & 24.5 & \textbf{526} & \textbf{12.9} \\
\bottomrule
\end{tabular}
\vspace{-10 pt}
\end{table}

\subsection{Implementation Details}
\label{sec: implementation}
All models are implemented in PyTorch~\cite{paszke2019pytorch} and trained on a single NVIDIA A100 GPU with a batch size of 16 for a total of 200k training steps, and all inference experiments are conducted on an NVIDIA RTX 4090 GPU. 
We adopt the CLIP-Base~\cite{radford2021learning} text and image encoders for language and global visual representation, and DINOv2-Base~\cite{oquab2023dinov2} as the dense visual backbone. 
Before full training, we first pretrain the model with only the anchor loss for 20k steps to establish a stable spatial attention ability. 
The FMT shares the same DiT-style architecture and hyperparameters as Diffusion Policy \cite{chi2025diffusion}, but differs in using Flow Matching for training.
The Gaussian width for the anchor region is set to $\sigma_{\text{task}} = \tfrac{1}{10}$ and $\sigma_{\text{end}}=\tfrac{1}{15}$ of the shorter side of the input image, and the anchor supervision weight is $\lambda{=}1.0$. $\sigma_{\text{end}}{<}\sigma_{\text{task}}$
For all methods, the action chunk length is fixed to 30.
Additional implementation details and the configurations of all baseline methods are provided in the Appendix.

\subsection{Evaluation}
\label{sec: result}
\paragraph{Main results.}
Table~\ref{tab:grasp_success} reports the grasp success rates of all evaluated methods under various testing conditions, including unseen backgrounds, lighting variations, distractor objects, and unseen objects. 
Overall, PosA-VLA achieves the highest success rates across nearly all scenarios, demonstrating strong generalization and robustness in visually diverse environments. 
In the standard \textit{Basic} setting, PosA-VLA reaches a success rate of 74.9\%, outperforming all baseline methods—including parameter-efficient variants such as $\pi_{0.5}$ (LoRA) and the full $\pi_{0.5}$—by a large margin. 
Under challenging conditions such as \textit{Unseen Background}, \textit{Unseen Lights}, and \textit{Distractor Objects}, PosA-VLA continues to show clear superiority, achieving 57.2\%, 47.8\%, and 56.1\% respectively—representing improvements of roughly 4--6\% over the strongest baseline (DexGraspVLA). 
These results highlight the effectiveness of our pose-conditioned supervision, which enables the model to maintain stable spatial attention and avoid distractions from irrelevant objects even under significant visual variations.

In the \textit{unseen object} setting, PosA-VLA performs worse than DexGraspVLA. 
This is primarily because DexGraspVLA leverages large-scale pretrained grounding and segmentation models to obtain precise object masks, providing stronger priors for novel object. 
Nevertheless, A key limitation of Dexgraspvla is its reliance on large pretrained models for detection and segmentation. 
When the perception module fails—e.g., incorrect grounding or missing object segmentation—the action policy also fails, regardless of how well the action model is trained. 
This dependence makes perception errors difficult to correct through downstream learning.
PosA-VLA still achieves competitive performance without relying on any external segmentation or foundation models, confirming its superior efficiency and generalizability for real-world manipulation tasks.

\paragraph{Long-horizon Manipulation.}
Table~\ref{tab:long_task} presents the results on the long-horizon manipulation task, where the robot must first open the box lid, then place the lid aside, and finally put a designated object inside the box. 
This task requires precise spatial reasoning, stable gripper control, and consistent action planning across multiple sequential stages. 
As shown in the table, PosA-VLA achieves the highest success rates in both sub-stages and the overall task, reaching 97.2\% for lid opening and placing, 63.0\% for pick up the object, and resulting in an overall success rate of 61.1\%. 
Compared with other VLAs, which often fail to maintain coherent attention across steps and produce unstable trajectories (e.g., OpenVLA-OFT and Smol-VLA both fall below 20\% overall success), PosA-VLA demonstrates strong spatial grounding and long-horizon consistency. 
It is also worth noting that DexGraspVLA cannot be directly evaluated on this task, as its detection-based pipeline requires a predefined object bbox for each manipulation step and thus cannot handle relational or spatial instructions such as “place the lid below” or “place the lid right”. 
These results highlight that the proposed anchor supervision effectively guides the model to attend to the correct regions throughout the entire manipulation process, enabling reliable multi-step execution in complex environments.

\paragraph{Efficiency Analysis}

Table~\ref{tab:efficiency} presents a comprehensive comparison of both training and inference efficiency across all evaluated methods. 
All reported results are averaged over multiple objects and repeated trials to ensure measurement reliability.
All models are implemented and evaluated in PyTorch, except for $\pi_0$, which is based on the JAX framework~\cite{jax2018github}. 
In terms of training cost, PosA-VLA requires only 20 GPU hours, comparable to Smol-VLA (19 GPU hours) and substantially lower than large-scale models such as OpenVLA-OFT (104 GPU hours) and DexGraspVLA (90 GPU hours). 
This demonstrates that our model can be trained efficiently without relying on heavy pretraining or complex architectural components. 
During inference, PosA-VLA achieves a favorable balance between speed and control stability, requiring only 24.5\,ms to execute a single action—faster than most baselines except Smol-VLA. 
It also reduces the average number of action steps to 526, the fewest among all compared methods. 
These results indicate that the proposed mechanism enables the policy to generate more decisive and accurate actions with fewer redundant motions. 
Consequently, PosA-VLA attains the shortest overall execution time of 12.9\,s, outperforming all baselines and demonstrating its suitability for real-time robotic manipulation with low computational overhead.

\begin{table}[t]
\centering
\caption{Ablation study on key components of PosA-VLA. 
We report the average grasp success rate (\%) after removing different modules.}
\label{tab:ablation}
\setlength{\tabcolsep}{3pt} 
\begin{tabular}{lc}
\toprule
\textbf{Model Variant} & \textbf{Average Success (\%)} \\
\midrule
PosA-VLA (full model)                   & \textbf{74.9} \\
w/o Anchor Loss                        & 29.5 \\
w/o Batch-wise Contrastive Loss       & 57.8 \\
w/o End-effector Attention                 & 55.6 \\
\bottomrule
\end{tabular}
\vspace{-5 pt}
\end{table}

\begin{table}[t]
\centering
\caption{Comparison of grasp success rates (\%) with different numbers of training samples per object. 
PosA-VLA achieves consistently higher performance even with limited data, showing strong training efficiency and robustness.}
\label{tab:sample_efficiency}
\setlength{\tabcolsep}{3pt}
\begin{tabular}{lcccc}
\toprule
\textbf{Method} & \textbf{50 samples} & \textbf{100 samples} & \textbf{200 samples} \\
\midrule
$\pi_0$              & 37.4 & 42.5 & 43.1 \\
OpenVLA\text{-}OFT   & 5.3  & 14.9 & 21.6 \\
Smol\text{-}VLA      & 0.0  & 18.7 & 20.3 \\
DexGraspVLA          & 40.1 & 49.2 & 57.2 \\
\midrule
\textbf{PosA-VLA} & \textbf{51.2} & \textbf{68.9} & \textbf{74.9} \\
\bottomrule
\end{tabular}
\vspace{-10 pt}
\end{table}

\subsection{Ablation study}
\label{sec: ablation}
\paragraph{Ablation on each module.}
\label{subsec:abl module}
We conduct an ablation study to evaluate the contribution of each key component in PosA-VLA, as summarized in Table~\ref{tab:ablation}. 
Removing the anchor supervision loss leads to a drastic drop in performance (29.5\%), confirming that the spatial supervision provided by the pose-conditioned attention regions is crucial for grounding the anchor attention weight $\mathbf{M}_t$. 
Without the batch-wise contrastive loss, the model fails to maintain cross-sample consistency in the joint text–image embedding space, resulting in fragmented and less discriminative attention features. 
Similarly, removing the end effector attention branch causes a notable decline to 55.6\%, as the model loses awareness of the end-effector’s position during manipulation, leading to unstable trajectories and inaccurate grasps. 
In contrast, the full PosA-VLA model achieves the highest average success rate, demonstrating that the combination of all modules effectively enhances spatial grounding and action consistency.

\paragraph{Ablation on Data Efficiency.}

To evaluate the data efficiency of PosA-VLA, we trained all models with different amounts of demonstration data and report the average grasp success rates in Table~\ref{tab:sample_efficiency}. 
When trained with only 50 samples, most baselines experience severe performance degradation, with OpenVLA and Smol-VLA almost failing to generalize (5.3\% and 0.0\% respectively), while DexGraspVLA achieves moderate success, which can be partly explained by its reliance on a large-scale foundation model~\cite{wang2024qwen2} that provides stronger visual grounding and language understanding.
In contrast, PosA-VLA maintains strong performance even under this low-data regime, achieving 51.2\% success, significantly higher than all baselines. 
With 100 training samples, PosA-VLA further improves to 68.9\%, showing faster learning progress and better scalability than other models. 
At the full 200 sample setting, PosA-VLA reaches 74.9\%, demonstrating stable convergence and high overall performance. 
These results highlight that the proposed anchor-based spatial supervision and multimodal alignment enable efficient learning from limited demonstrations, making PosA-VLA more practical for real-world robotic deployment where annotated data are scarce.

%% file: sec/5_conclusion.tex
\section{Conclusion}
In this work, we present \textbf{PosA-VLA}, a vision-language-action framework that anchors attention via pose-conditioned supervision to improve spatial grounding and control consistency in robotic manipulation. 
By introducing pose-conditioned attention anchors derived from the robot’s end-effector trajectories, our method explicitly guides the model to attend to both task-relevant regions and end-effector poses throughout the manipulation process. 
A batch-wise contrastive objective further enhances cross-sample consistency in multimodal alignment, while the integration of a Flow Matching Transformer enables efficient and smooth action generation. 
Comprehensive experiments demonstrate that PosA-VLA achieves superior performance across diverse robotic manipulation benchmarks, exhibiting stronger spatial awareness, faster inference, and more stable long-horizon control compared with existing VLAs. 
In the future, we plan to extend our approach to more complex manipulation scenes involving multiple objects and dynamic environments, and to explore integrating large-scale world models for open-vocabulary, goal-conditioned robotic behavior, which are still underexplored directions in the field.

%% file: sec/X_suppl.tex
\clearpage
\setcounter{page}{1}
\maketitlesupplementary

\section{Implementation details}
\label{supsec:implementation}

\begin{table}[h]
\centering
\caption{Implementation details and hyperparameters used for PosA-VLA training and inference.}
\label{tab:impl_detail}
\setlength{\tabcolsep}{0.5pt}
\begin{tabular}{lc}
\toprule
\textbf{Hyperparameter} & \textbf{Value} \\
\midrule
Learning rate & $3.0\times10^{-4}$ \\
Optimizer & AdamW~\cite{loshchilov2017decoupled} \\
AdamW betas & $[0.95,\ 0.999]$ \\
AdamW $\epsilon$ & $1.0\times10^{-8}$ \\
Weight decay & $1.0\times10^{-6}$ \\
LR scheduler & Cosine annealing \\
Warmup steps & 2000 \\
Gradient accumulation & 1 \\
Flow-matching time sampling & $\text{Beta}(\alpha{=}1.5,\ \beta{=}1.0)$ \\
Time range & $[0.001,\ 1.0]$ \\
Flow-matching inference steps ($N_{\text{infer}}$) & 10 \\
\bottomrule
\end{tabular}
\end{table}

We summarize the detailed hyperparameter settings of \textbf{PosA-VLA} in Table~\ref{tab:impl_detail}. 

For fair comparison, the hyperparameters of baseline methods follow their original implementations, except for the total training steps, batch size, and number of GPUs used. 
Specifically:  
\begin{itemize}
    \item \textbf{DexGraspVLA}~\cite{zhong2025dexgraspvla}: trained for 60 epochs with a batch size of 36 on 8~GPUs.  
    \item \textbf{$\pi_0$}~\cite{black2410pi0}: trained for 30{,}000 steps with a batch size of 32 on a single GPU.  
    \item \textbf{Smol-VLA}~\cite{shukor2025smolvla}: trained for 200{,}000 steps with a batch size of 16 on a single GPU.  
    \item \textbf{OpenVLA-OFT}~\cite{kim2025fine}: trained for 100{,}000 steps with a batch size of 8 on 8~GPUs.  
\end{itemize}

All remaining hyperparameters, including learning rate, optimizer settings, and data preprocessing, follow the configurations reported in the respective original papers.

\section{Training Data Setup}
\begin{figure}[t]
    \centering
    \includegraphics[width=0.9\linewidth]{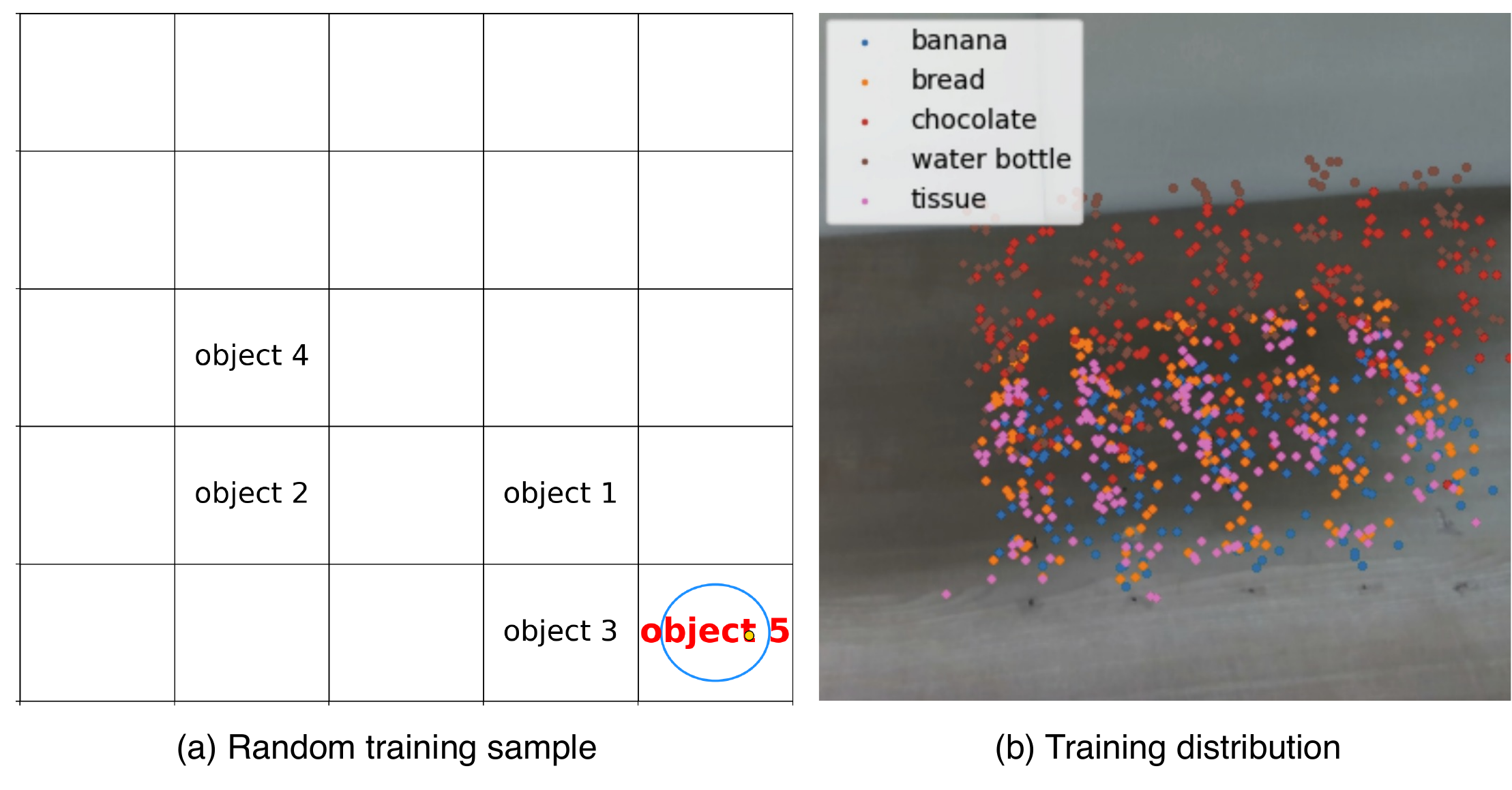}
    \caption{    (a) Illustration of our data collection setup. 
    Objects are randomly placed on a $5{\times}5$ grid, where the red object denotes the target to be grasped, and the yellow dots on the circle indicate the reference points for human placement. 
    (b) Visualization of the overall training distribution obtained by projecting all grasp points from the demonstrations back onto the 2D image plane.
}
\label{demo: training distribution}
\vspace{-10 pt}
\end{figure}

\begin{figure}[t]
    \centering
    \includegraphics[width=0.9\linewidth]{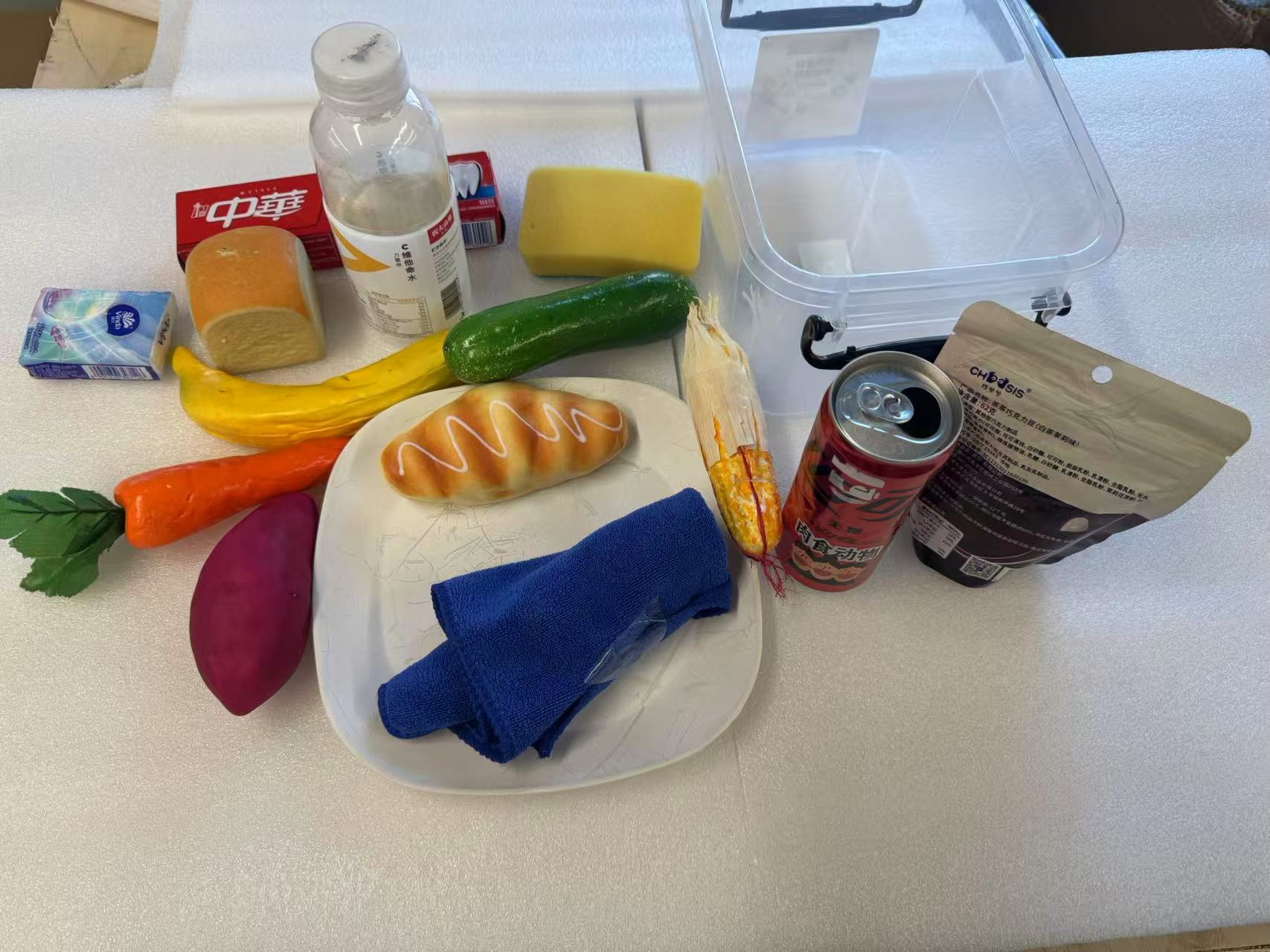}
    \caption{All physical objects and props used in our real-world experiments. 
The set includes various household items such as a water bottle, bread, carrot, tissue bag, chocolate bag, and a box with a detachable lid, which are used across different tasks including grasping, placing, and long-horizon manipulations. 
These objects cover diverse shapes, textures, and materials, providing rich visual and physical variability for evaluating the generalization capability of PosA-VLA.
}
\label{demo: objects}
\vspace{-10 pt}
\end{figure}

To ensure that the collected demonstrations cover a diverse range of action distributions while maintaining full coverage of the robot’s reachable workspace and camera field of view, we discretize the tabletop into a $5{\times}5$ grid of \emph{25} center points. 
For each target object, we iterate over all centers and collect demonstrations via teleoperation under randomized placements. 
Concretely, at each center $c_i$, the target object is placed within a small disk of radius $r$ around $c_i$ (uniformly sampled translation and in-plane rotation), ensuring it lies within both the arm’s reachable region and the cameras’ FOV. 
To introduce clutter and distractors, we randomly sample \emph{four} additional grid centers per trial and place non-target objects around those centers using the same within-disk perturbation. 
For every center, we record \emph{8} teleoperated grasp trials, yielding $25\times 8=200$ demonstrations per object. 
Across all trials, we randomize the target’s in-plane pose and the distractors’ positions to induce variability while avoiding collisions and excessive occlusions. 
An example of a sampled demonstration is shown in Figure~\ref{demo: training distribution}(a), where objects are randomly distributed over the $5{\times}5$ grid and the red object indicates the target to be grasped. 
Furthermore, we project all grasp points from the training demonstrations of different objects back onto the 2D image plane to visualize the overall training distribution, as illustrated in Figure~\ref{demo: training distribution}(b). 
This protocol produces balanced spatial coverage and consistent task difficulty while enriching the action diversity in the dataset, resulting in a total of 200 demonstrations per object for training. 
Each demonstration is represented as a sequence of continuous robot actions, including the Cartesian position $(x,y,z)$, the 6D end-effector orientation~\cite{zhou2019continuity}, the joint rotation angles, and the gripper open/close state. 
All physical objects and props used in the experiments are shown in Figure~\ref{demo: objects}.

\section{Evaluation Details}
Our evaluation environments are illustrated in Figure~4 of the main paper. 
To ensure statistical reliability and eliminate potential sampling bias, we evaluate all grasping tasks by randomly placing each target object \textbf{100 times} within the workspace. 
For each trial, four additional distractor objects are also randomly positioned on the tabletop to introduce clutter and interference. 
All target and distractor placements are guaranteed to be \emph{unseen} during training, ensuring that the model is evaluated on novel configurations rather than memorized layouts. 

For the \textit{Long-horizon} task, both the box and the target object are randomly positioned at the beginning of each episode, again avoiding any overlap with training positions. 
Each target object designated to be placed into the box is tested \textbf{50 times}, providing sufficient coverage for evaluating consistency and robustness across varied spatial arrangements. 
This evaluation protocol guarantees fair comparison among all methods and reflects realistic deployment conditions in unseen, cluttered environments.

\section{Simulation experiments}
We further evaluate the generalization ability of PosA-VLA in simulated environments using the \textbf{Libero}~\cite{liu2023libero} benchmark, which consists of multiple vision-language-conditioned manipulation tasks categorized into \textit{Spatial}, \textit{Object}, and \textit{Long-horizon} settings. 
To adapt our model to simulation, we use the CLIP image encoder as the visual backbone instead of DINOv2, since we observed that DINOv2 features underperform on Libero’s synthetic renderings. 
The CLIP visual features are directly fed into the Flow Matching Transformer (FMT) for action prediction, while all other architectural and training configurations remain identical to those used in real-world experiments.

Table~\ref{tab:libero} summarizes the quantitative results. 
PosA-VLA achieves competitive or superior performance across all three task categories, with an overall average success rate of 95.1\%, comparable to or exceeding the best-performing baselines. 
Compared with methods trained on modified datasets (marked with~*), which filter out unsuccessful demonstrations, PosA-VLA attains comparable accuracy without requiring such dataset curation, highlighting its robustness and data efficiency. 
These results confirm that the proposed pose-conditioned anchor attention effectively transfers to simulation, enabling strong performance under diverse visual and physical settings.

\begin{table}[t]
\centering
\caption{
Comparison of performance on the \textbf{Libero} simulation benchmark.
We report average success rates (\%) across multiple manipulation tasks. * means the trained on the modified training dataset with unsuccessful demonstrations filtered out
}
\label{tab:libero}
\setlength{\tabcolsep}{4pt}
\begin{tabular}{lcccc}
\toprule
\textbf{Method} & \textbf{Spatial} & \textbf{Object} & \textbf{Long} & \textbf{Average} \\
\midrule
Smol-VLA             & 93.0 & 94.0 & 77.0 & 88.0 \\
OpenVLA-OFT          & 95.2 & 94.2 & 93.2 & 94.2 \\
$\pi_0$-FAST*              & 96.4 & 96.8 & 60.2 & 84.5 \\
$\pi_0$*              & \textbf{96.8} & \textbf{98.8} & 85.2 & 93.6 \\
OpenVLA-OFT*          & 97.6 & 98.4 & \textbf{94.5} & \textbf{96.8} \\
\midrule
\textbf{PosA-VLA (ours)} & 95.6 & 98.6 & 91.2 & 95.1 \\
\bottomrule
\end{tabular}
\vspace{-6pt}
\end{table}

\section{Pseudo Code for Anchor Loss Training}
To clearly illustrate how the pose-conditioned anchor loss is optimized, we provide a concise pseudo-code implementation in Algorithm~\ref{alg:focus_loss_short}. 
The algorithm describes the forward and supervision pipeline for computing the spatial attention loss and batch-wise contrastive loss.
Specifically, it shows how task-relevant and end-effector anchor maps are generated from projected gripper poses, how positive indices are selected according to activation thresholds, and how both losses are combined to form the total anchor loss $\mathcal{L}_{\text{anchor}}$. 
This pseudo code offers a clear overview of the optimization process used to align the predicted anchor attention weights with spatial supervision during training.



\begin{algorithm}[t]
\caption{Pose-Conditioned Anchor Loss Training for PosA-VLA}
\label{alg:focus_loss_short}
\SetAlgoLined
\KwIn{Dataset $\mathcal{D}=\{(\mathbf{x}, \mathbf{I}^h_t, \mathbf{I}^w_t, \mathbf{p}_t)\}$, parameters $\theta$, weight $\alpha$.}
\KwOut{Updated model parameters $\theta$.}

\For{each mini-batch $(\mathbf{x}, \mathbf{I}^h_t, \mathbf{I}^w_t, \mathbf{p}_t)$ in $\mathcal{D}$}{
    Extract text embedding $\mathbf{f}_x$ (instruction) and $\mathbf{f}_e$ (end-effector query, e.g., ``gripper'') using the CLIP text encoder. \\
    Extract patch-wise visual embeddings $\mathbf{F}_I$ from both cameras using the CLIP image encoder. \\
    Fuse $\mathbf{f}_x$ with $\mathbf{F}_I$ via cross-attention to predict task-relevant attention $\mathbf{M}_t^{\text{task}}$. \\
    Fuse $\mathbf{f}_e$ with $\mathbf{F}_I$ via cross-attention to predict end-effector attention $\mathbf{M}_t^{\text{end}}$. \\
    Stack to form pose-anchored attention weights: $\mathbf{M}_t = [\mathbf{M}_t^{\text{task}}, \mathbf{M}_t^{\text{end}}]$. \\
    Project 3D end-effector pose $\mathbf{p}_t$ onto 2D coordinates $(u_t, v_t)$ using the camera–end-effector transformation. \\
    Generate Gaussian supervision maps $\mathbf{F}_f^{\text{task}}$ and $\mathbf{F}_f^{\text{end}}$ centered at $(u_t, v_t)$ with different variances $\sigma_{\text{task}}$ and $\sigma_{\text{end}}$. \\
    Compute spatial loss: $\mathcal{L}_{\text{f}} = \text{FocalLoss}(\mathbf{M}_t, [\mathbf{F}_f^{\text{task}}, \mathbf{F}_f^{\text{end}}])$. \\
    Select positive indices $\Omega^+ = \{(i,j)\mid \mathbf{F}_f(i,j) > 0.7\}$ and extract visual embeddings $\mathbf{v}_{\text{obj}}, \mathbf{v}_{\text{end}}$ from $\mathbf{F}_I$. \\
    Form fused multimodal embeddings $\mathbf{z} = [\mathbf{f}_x; \mathbf{f}_e; \mathbf{v}_{\text{end}}; \mathbf{v}_{\text{obj}}]$. \\
    Compute batch-wise contrastive loss $\mathcal{L}_{\text{c}}$ using pairwise BCE over all $(m,n)\!\in\!\mathcal{P}$. \\
    Combine: $\mathcal{L}_{\text{anchor}} = \alpha\,\mathcal{L}_{\text{f}} + (1-\alpha)\mathcal{L}_{\text{c}}$. \\
    Update model parameters: $\theta \leftarrow \theta - \eta \nabla_\theta \mathcal{L}_{\text{anchor}}$. \\
}
\end{algorithm}

\section{Additional Real-World Experiments}
To further validate the effectiveness of our pose-conditioned anchor attention, we design two additional types of real-world tasks that require spatial reasoning with respect to the head camera’s coordinate frame. 
The first task involves grasping a piece of bread and placing it to the \textit{left}, \textit{right}, or \textit{top} side of the scene based on the head camera view (Figure~\ref{demo:bread}). 
The second task requires grasping either the \textit{lower end} or the \textit{upper end} of a carrot and placing it on the \textit{left} or \textit{right} side of a plate (Figure~\ref{demo:carrot}). 
In both tasks, PosA-VLA accurately anchors attention to both the object being manipulated and the target placement region, demonstrating robust spatial awareness and precise goal grounding. 
These results highlight that our anchor attention not only enables reliable localization of the manipulated object but also effectively anchors to the intended goal position, allowing successful task completion under complex spatial relationships.

\section{Visualization Results}
\begin{figure*}[t]
    \centering
    \includegraphics[width=0.9\linewidth]{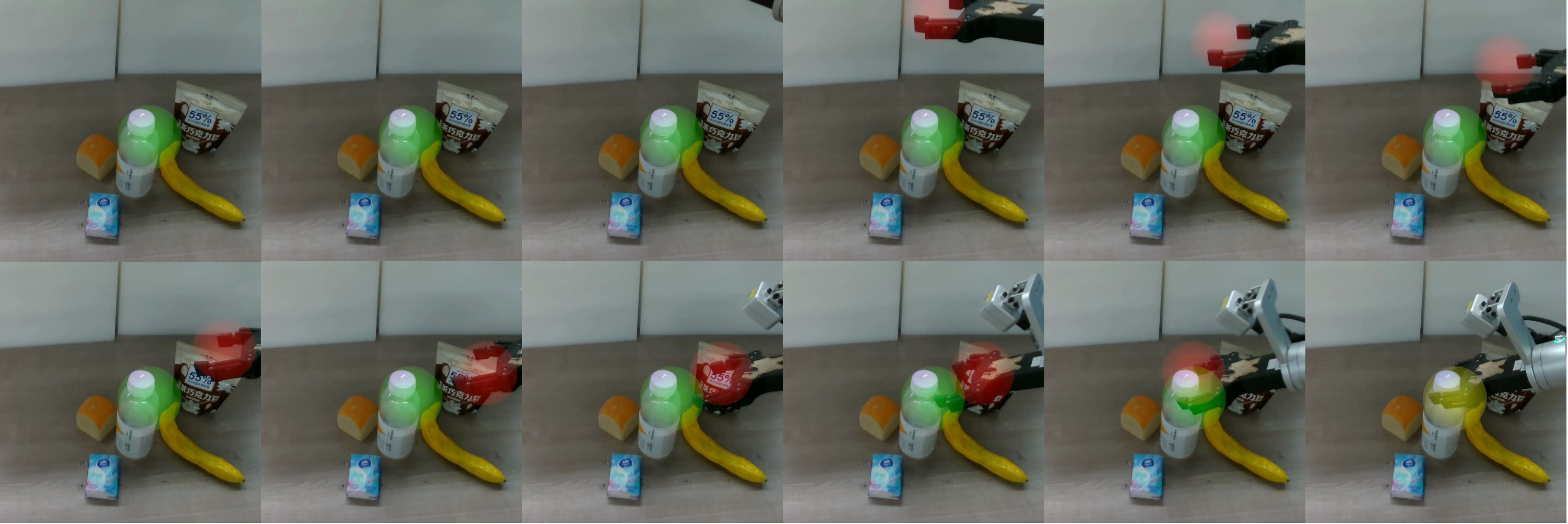}
    \caption{Grasping demonstration with \textit{anchor-attention weight} visualization. Task: pick up the water bottle. Green circle represents task-relevant attention, while red circle represents end-effector attention. The combination of them is yellow circle.
}
\label{demo:attention}
\vspace{-10 pt}
\end{figure*}

\begin{figure*}[t]
    \centering
    \includegraphics[width=0.9\linewidth]{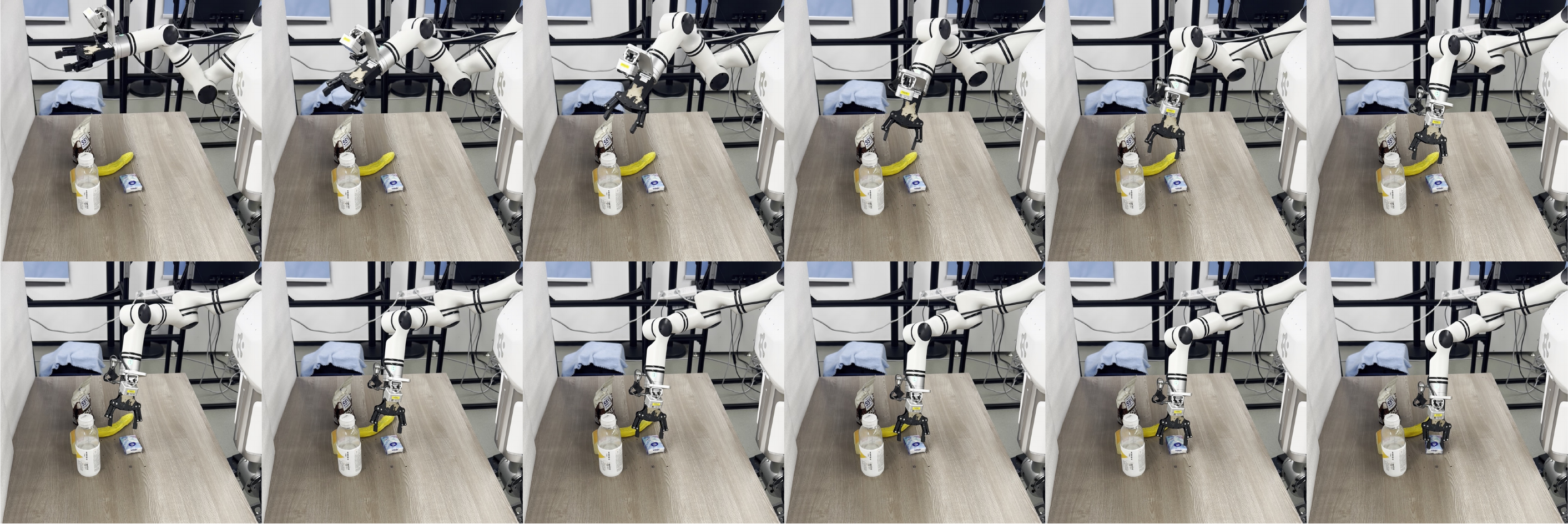}
    \caption{Grasping demonstration under the \textit{Basic} setting. Task: pick up the tissue bag.
}
\label{demo:basic}
\vspace{-10 pt}
\end{figure*}

\begin{figure*}[t]
    \centering
    \includegraphics[width=0.9\linewidth]{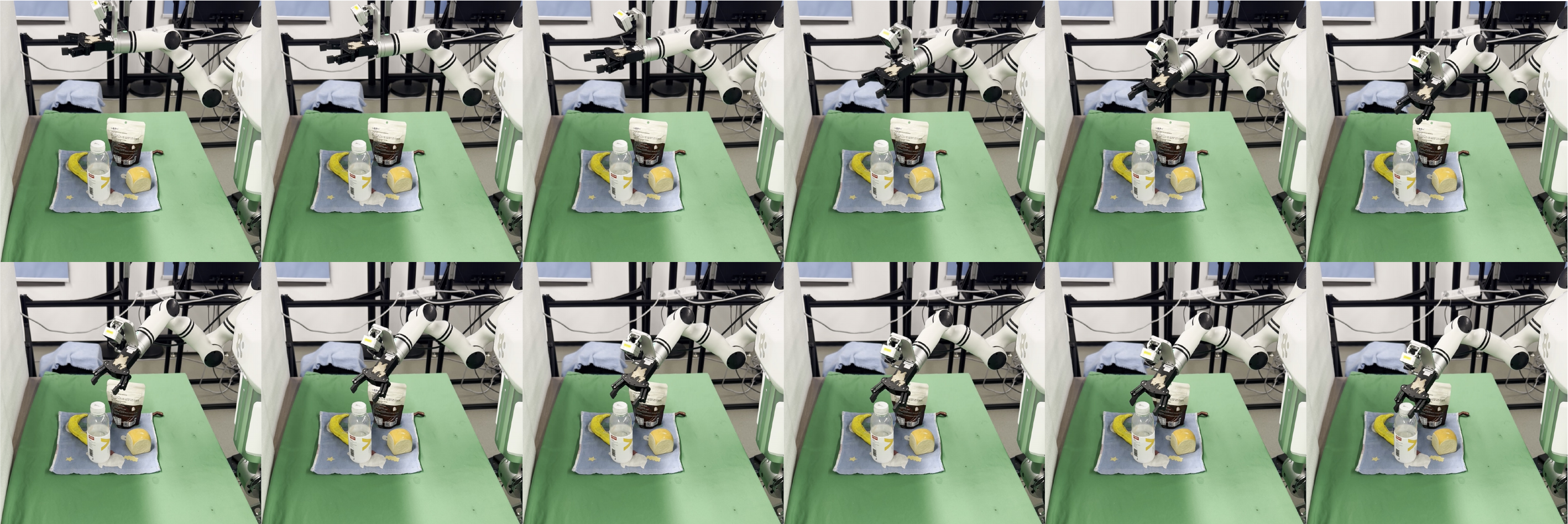}
    \caption{Grasping demonstration under the \textit{Unseen Background} setting. Task: pick up the water bottle.
}
\label{demo:background}
\vspace{-10 pt}
\end{figure*}

\begin{figure*}[t]
    \centering
    \includegraphics[width=0.9\linewidth]{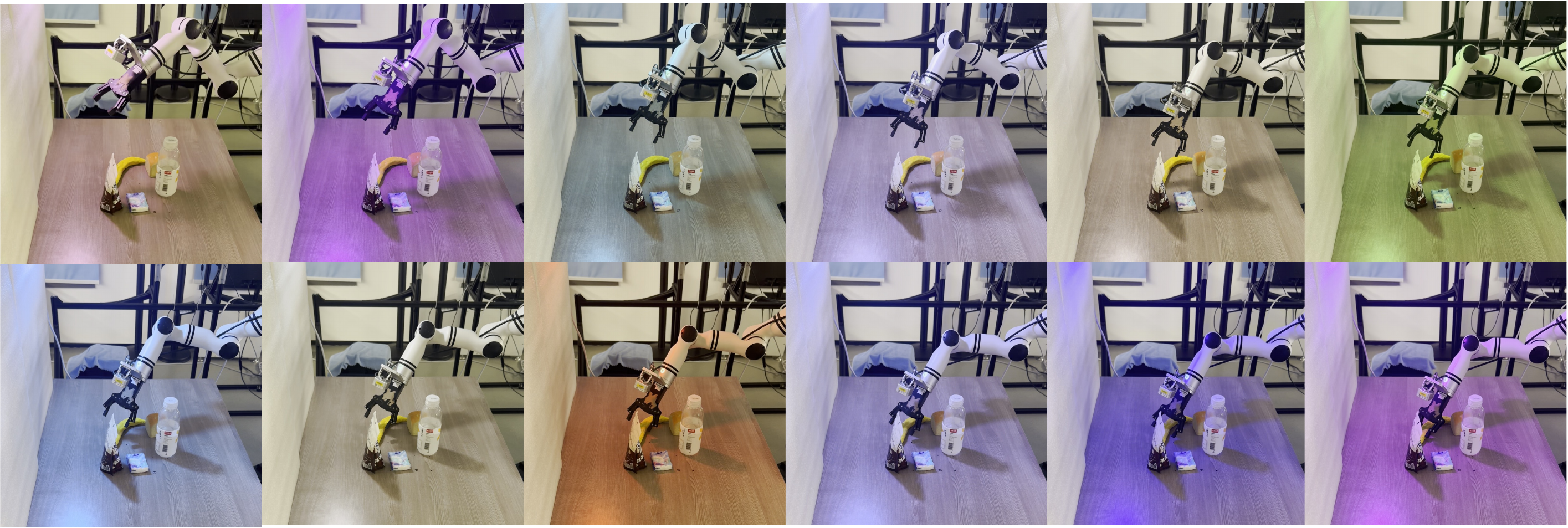}
    \caption{Grasping demonstration under the \textit{Unseen Lighting} condition. Task: pick up the chocolate bag.
}
\label{demo:light}
\vspace{-10 pt}
\end{figure*}

\begin{figure*}[t]
    \centering
    \includegraphics[width=0.9\linewidth]{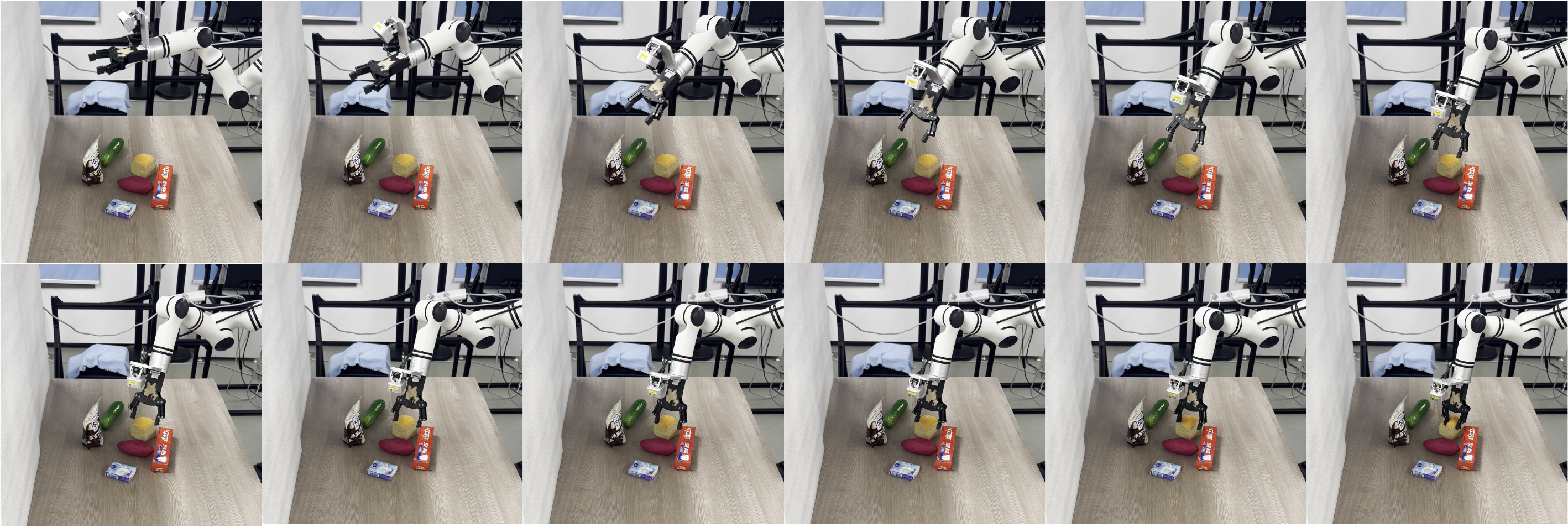}
    \caption{Grasping demonstration in the presence of \textit{Distractor Objects}. Task: pick up the bread.
}
\label{demo:distractor}
\vspace{-10 pt}
\end{figure*}

\begin{figure*}[t]
    \centering
    \includegraphics[width=0.9\linewidth]{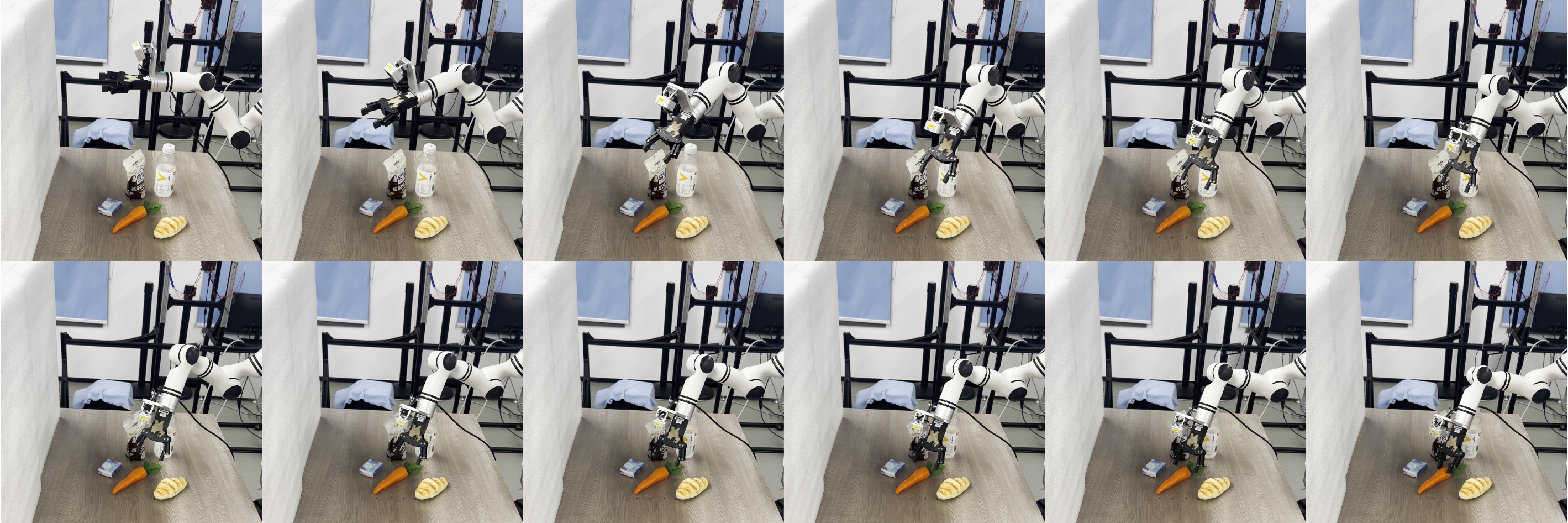}
    \caption{Grasping demonstration with an \textit{Unseen Object}. Task: pick up the carrot.
}
\label{demo}
\vspace{-10 pt}
\end{figure*}

\begin{figure*}[t]
    \centering
    \includegraphics[width=0.9\linewidth]{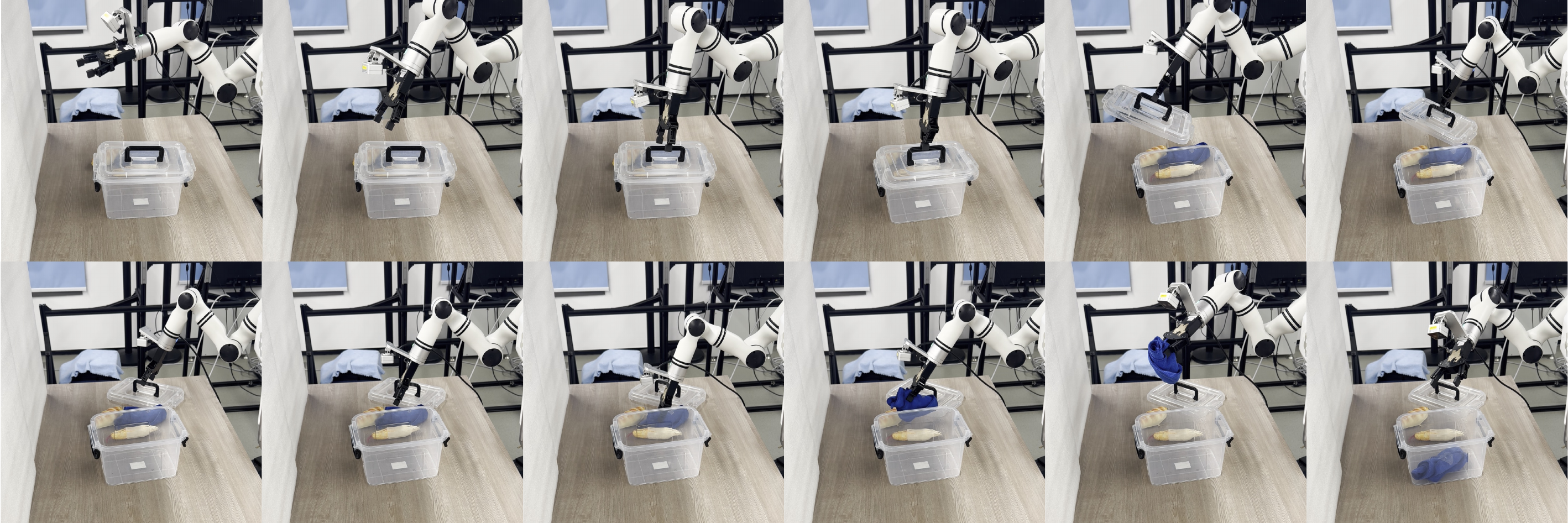}
    \caption{Long-horizon manipulation task where the robot opens the box lid, put the box lid aside and places a designated object inside the box. 
}
\label{demo:long}
\vspace{-10 pt}
\end{figure*}

\begin{figure*}[t]
    \centering
    \includegraphics[width=0.9\linewidth]{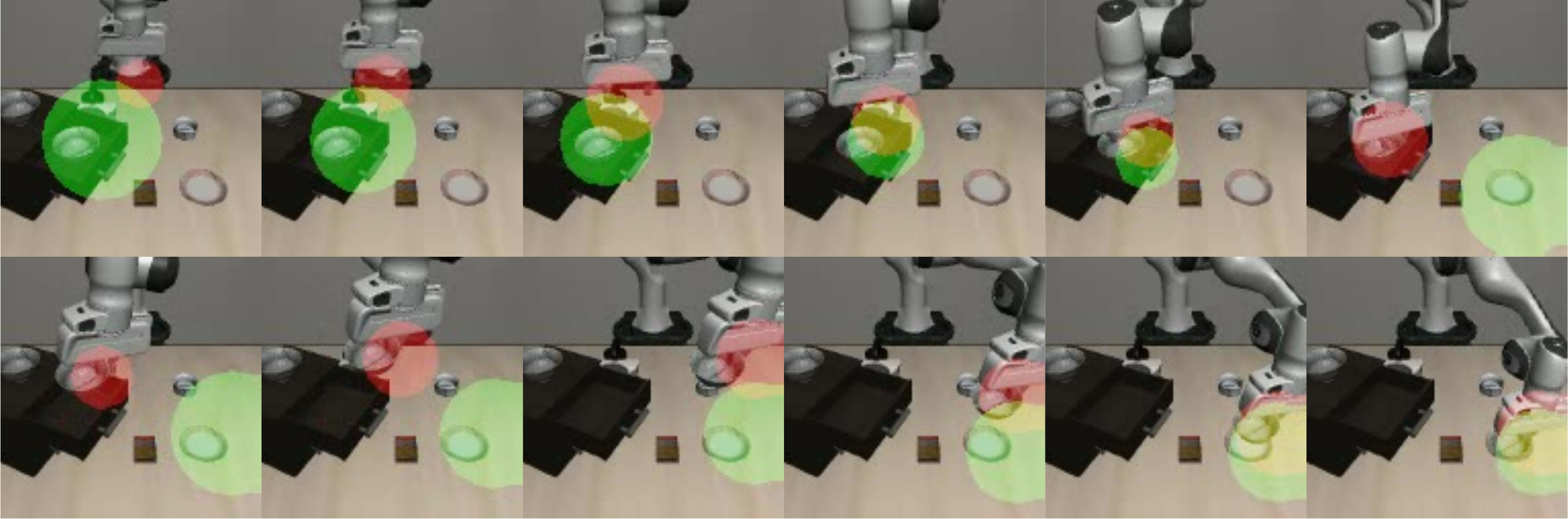}
    \caption{Simulation demonstration on the \textbf{Libero}~\cite{liu2023libero} benchmark with anchor attention. 
}
\label{demo:libero}
\vspace{-10 pt}
\end{figure*}

\begin{figure*}[t]
    \centering
    \includegraphics[width=0.9\linewidth]{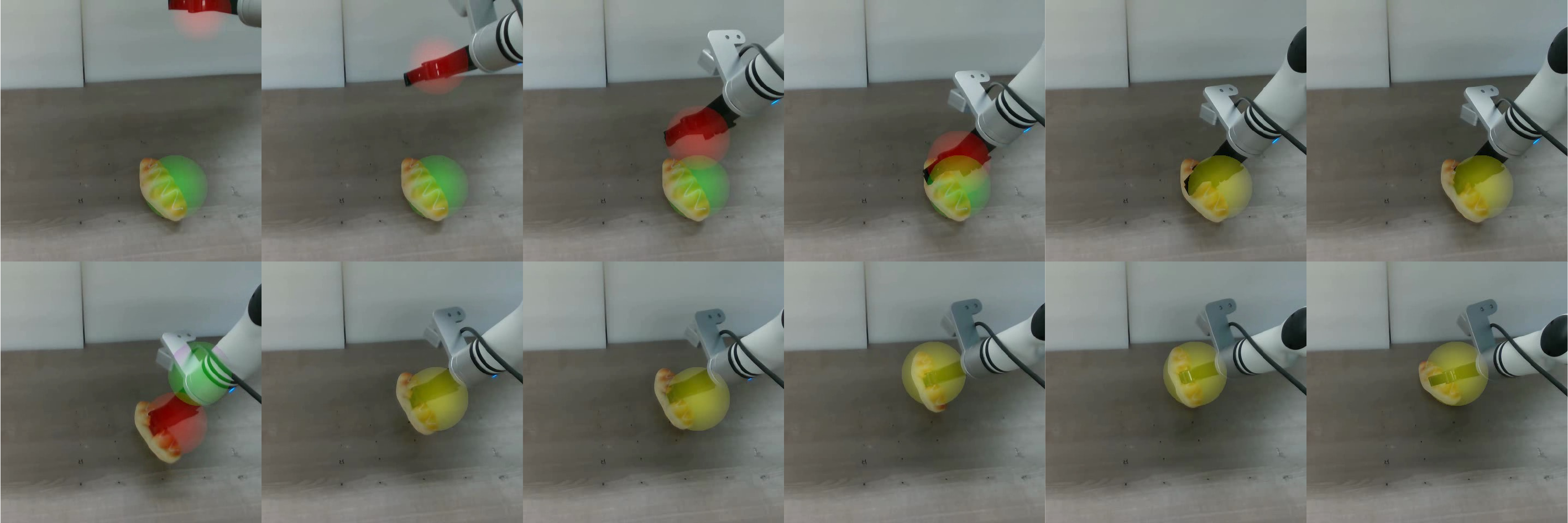}
    \caption{Demonstrations of newly defined composite tasks: grasping a piece of bread and placing it on the top side of the table. Green circle represents task-relevant attention, while red circle represents end-effector attention. The combination of them is yellow circle.
}
\label{demo:bread}
\vspace{-10 pt}
\end{figure*}

\begin{figure*}[t]
    \centering
    \includegraphics[width=0.9\linewidth]{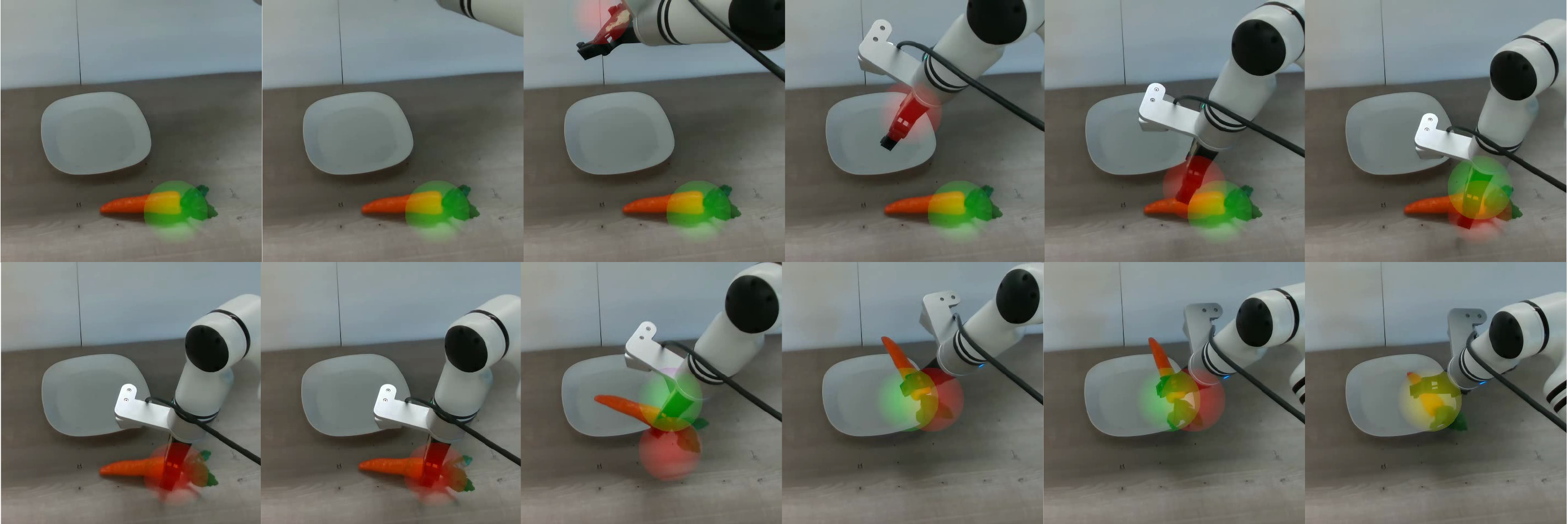}
    \caption{Demonstrations of newly defined composite tasks: grasping the lower end of a carrot and placing it on the right side of a plate. Green circle represents task-relevant attention, while red circle represents end-effector attention. The combination of them is yellow circle.
}
\label{demo:carrot}
\vspace{-10 pt}
\end{figure*}

We provide extensive qualitative visualizations to further demonstrate the performance and interpretability of \textbf{PosA-VLA} in both real-world and simulated environments. 
Figures~\ref{demo:attention}–\ref{demo:carrot} present representative grasping and manipulation sequences across various testing conditions, including \textit{Basic}, \textit{Unseen Background}, \textit{Unseen Lighting}, \textit{Distractor Objects}, \textit{Unseen Objects}, and \textit{Long-horizon Tasks}, as well as simulation results on the \textbf{Libero} benchmark. 
In addition, we visualize anchor-attention weights to show how the model dynamically attends to task-relevant regions (green) and end-effector positions (red), whose combination forms the final attention anchor (yellow). 
The additional composite tasks—such as grasping and relocating a piece of bread or manipulating the lower end of a carrot—further verify that our model can accurately anchor both the object and the target placement area to complete complex goal-directed actions. 
For better visualization of dynamic motion and attention evolution, please refer to the accompanying video demonstrations provided in the attached \texttt{.zip} archive.

\section{Limitations}
While PosA-VLA demonstrates strong performance across various manipulation scenarios, it still has several limitations. 
First, our method relies on end-effector state changes (e.g., gripper opening or closing) to generate pose-conditioned anchors. 
As a result, it is less effective for actions where the end-effector maintains a fixed configuration, such as pushing or sliding objects across the surface. 
Second, PosA-VLA assumes clear visual observation of the target during supervision. 
When the target object is heavily occluded, the focus-region supervision derived from projected gripper positions may become inaccurate, leading to misaligned attention and degraded performance. 
Addressing these limitations—such as incorporating force or tactile cues and learning occlusion-aware spatial representations—will be an important direction for future work.

